\def\year{2022}\relax
\documentclass[letterpaper]{article} 
\usepackage{aaai22}  
\usepackage{times}  
\usepackage{helvet}  
\usepackage{courier}  
\usepackage[hyphens]{url}  
\usepackage{graphicx} 
\usepackage{natbib}  
\usepackage{caption} 
\DeclareCaptionStyle{ruled}{labelfont=normalfont,labelsep=colon,strut=off} 
\usepackage{algorithm}
\usepackage{algorithmic}

\usepackage{newfloat}
\usepackage{listings}
\floatstyle{ruled}
\newfloat{listing}{tb}{lst}{}
\floatname{listing}{Listing}
\usepackage{amsmath}
\usepackage{amsfonts}
\usepackage{graphicx}
\usepackage{xcolor}
\usepackage{kotex}
\usepackage{multirow}
\usepackage{svg}
\usepackage{amsmath}
\usepackage{array}
\usepackage{booktabs}
\usepackage{dcolumn}
\usepackage{caption}
\usepackage{subcaption}
\usepackage[multiple]{footmisc}

\usepackage{tikz}

\newcommand{\figref}[1]{Fig.~\ref{#1}}

\newcommand{\tabref}[1]{Table~\ref{#1}}

\newcommand{\Figref}[1]{Figure~\ref{#1}}

\title{CG-NeRF: Conditional Generative Neural Radiance Fields}
\author{Kyungmin Jo\thanks{Equal contribution}, Gyumin Shim\footnotemark[1], Sanghun Jung, Soyoung Yang, Jaegul Choo}

\affiliations{
    KAIST\\

    Graduate school of AI, Korea Advanced Institute of Science and Technology, Daejeon, Republic of Korea \\
    \texttt{\{bttkm, shimgyumin, shjung13, sy\_yang, jchoo\}@kaist.ac.kr}
}

\usepackage{bibentry}

\begin{document}

\makeatletter
\g@addto@macro\@maketitle{
  \begin{figure}[H]
  \captionsetup{width=0.95\textwidth}
  \setlength{\linewidth}{0.95\textwidth}
  \setlength{\hsize}{\textwidth}
  \centering
  \includegraphics[width=1\linewidth]{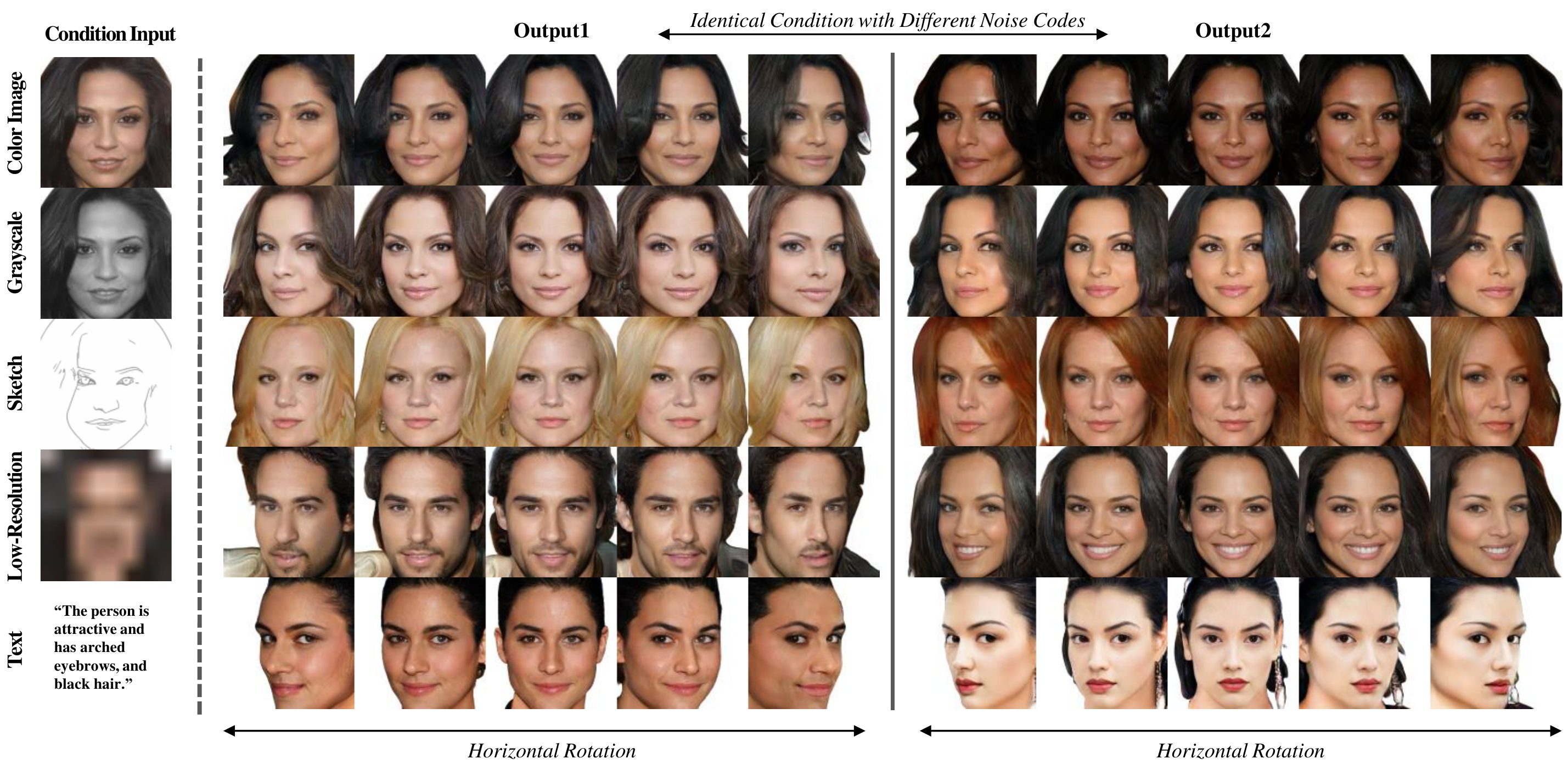} 
  \caption{\small Our method produces diverse 3D-aware output images reflecting various condition inputs (first column). For each condition input, two different output images generated with different noise codes are visualized with horizontal rotation.}
  \label{fig:teaser}
  \end{figure}
}
\maketitle

\begin{abstract}


While recent NeRF-based generative models achieve the generation of diverse 3D-aware images, these approaches have limitations when generating images that contain user-specified characteristics. In this paper, we propose a novel model, referred to as the conditional generative neural radiance fields (CG-NeRF), which can generate multi-view images reflecting extra input conditions such as images or texts. While preserving the common characteristics of a given input condition, the proposed model generates diverse images in fine detail. We propose: 1) a novel unified architecture which disentangles the shape and appearance from a condition given in various forms and 2) the pose-consistent diversity loss for generating multimodal outputs while maintaining consistency of the view. Experimental results show that the proposed method maintains consistent image quality on various condition types and achieves superior fidelity and diversity compared to existing NeRF-based generative models.



\section{Introduction}

The neural radiance field (NeRF)~\cite{mildenhall2020nerf} successfully addresses unseen view synthesis, a long-lasting problem in computer vision, by learning to construct 3D scene from a set of images taken from multiple viewpoints via a differentiable rendering technique.
%
Because NeRF takes the 3D coordinate and the viewpoint of a target scene as inputs, 
it is capable of synthesizing view-consistent images (\textit{i.e.,} images corresponding to the input view point).
Due to the success of NeRF, this approach has been widely extended to various fields, such as video ~\cite{li2021neural, xian2021space}, pose estimation~\cite{su2021nerf}, scene labeling and understanding ~\cite{zhi2021place}, and object category modeling~\cite{xie2021fig}.

While these techniques utilize NeRF only for synthesizing an unseen view, recent studies have emerged that 
generate photorealistic multi-view images based on generative adversarial networks(GANs)~\cite{GRAF, GIRAFFE, pigan}.
Compared to the existing 2D-based generative models, these studies can produce 3D-aware images by generating view-consistent images for given camera poses.
However, because the generative models produce images without any 
user-specified condition, these studies have limits regarding the generation of images that contain the desired characteristics of the condition, as shown in \figref{fig:teaser}.

To extend the task of existing generative NeRF models, we propose a novel task, termed conditional generative NeRF (CG-NeRF), which performs 3D-aware image synthesis for a given condition.
The proposed task aims to create view-consistent images with \emph{diverse} styles in fine detail while reflecting the characteristics of \emph{conditions}.
To the best of our knowledge, we newly tackle this task, extending existing generative NeRF approaches that do not consider user-specified conditions.
First, while aiming to reflect user-specified conditions, we realize that conditions can exist in various forms (\textit{e.g.,} texts and images). Thus, to accommodate such user-specified conditions, we propose a unified method adaptively applicable to various condition types, as processing those various types of conditions in an unified model can significantly enhance its applicability for users.
We consider various condition types, including color images, grayscale, sketches, low-resolution images, and text, as shown in the condition inputs in \figref{fig:teaser}.
To support these various type of conditions, we leverage a pre-trained semantic multimodal encoder, CLIP, and this is followed by the disentanglement of shape and appearance from the encoded vector of the input condition.

Generating diverse images is also one of the important factors. To produce various images, we design a generative model, capable of generating fine details while reflecting the common characteristics of the input conditions.
This is a challenging problem because fully relying on the input conditions may be an overly simple solution in terms of the generation model, and this problem could degrade the diversity.
To enhance the diversity of the output images further, we propose a novel pose-consistent diversity (PD) loss that explicitly penalizes view inconsistencies.
In summary, our contributions are as follows:

\begin{itemize}
\setlength\itemsep{0em}
    \item We newly define a novel task, the conditional generative neural radiance fields referred to as CG-NeRF.
    \item The proposed method generates diverse and photorealistic images reflecting the condition inputs, effectively disentangling the shape and appearance from the input condition.
    \item For improved diversity of the output images, we propose the pose-consistent diveristy (PD) loss, which helps to maximize only style differences while maintaining consistency of the view. 
    \item We conduct extensive experiments and demonstrate that our unified model can generate diverse images, reflecting various conditions.
    
\end{itemize}


\begin{figure*}[h!]
\centering
\begin{tabular}{@{}c}
\includegraphics[width=0.87\linewidth]{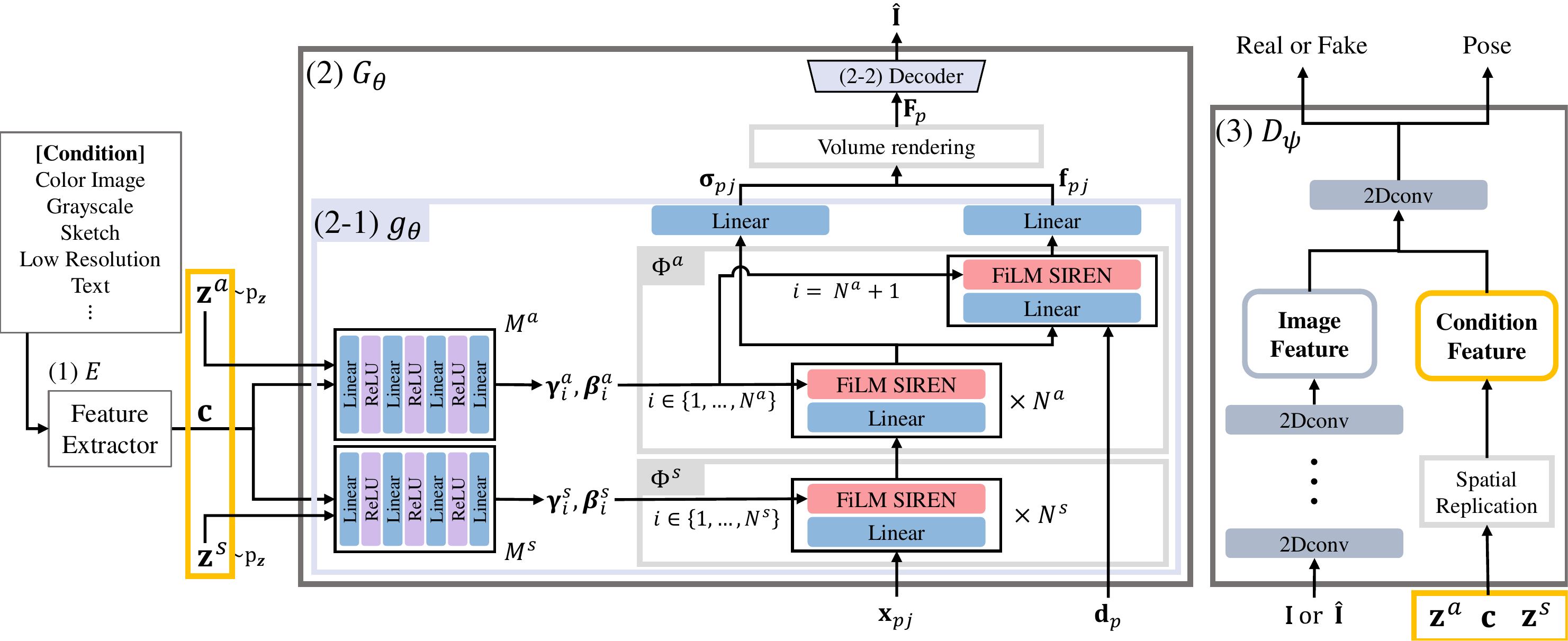} \\
\end{tabular}
\caption{\small Illustration of our main architecture. Notations are summarized in \tabref{tb:notation}.}
\label{fig:framework}
\end{figure*}

\section{Related Work}

\subsubsection{Neural Radiance Fields}
Recent advancements~\cite{mildenhall2020nerf, bakingNeRF, dietNeRF, nerv, pixelNeRF, autoint} in the area of novel view synthesis have been accomplished by employing the NeRF. The seminal work~\cite{mildenhall2020nerf} has proven the effectiveness of volume rendering with NeRF, and later studies~\cite{bakingNeRF, nerf--, nerf++} proposed further improvements over the original NeRF. While some NeRF studies enhance the original NeRF in terms of both quality and efficiency, our work is more related to generative NeRF methods, which have attracted attention recently.


\subsubsection{Generative NeRF}
Along with the improvements to the NeRF itself, generative NeRF models~\cite{GRAF, GIRAFFE, pigan, campari} have also emerged. GRAF~\cite{GRAF} proposes a generative model with implicit radiance fields for the novel scene generation. Moreover, GIRAFFE~\cite{GIRAFFE} improves GRAF by separating the object instances in a controllable way, which lets users gain more ability to compose new scenes. Another study, pi-GAN~\cite{pigan}, which is more closely related to our work, employs the SIREN~\cite{sitzmann2020implicit} activation function along with the multilayer perceptron (MLP), which is effective when used for novel scene generation. However, none of the previous approaches have attempted to add conditions while generating, though this would allow users to generate various scenes according to diverse conditions. Therefore, in this work, we propose a novel model, CG-NeRF, which can significantly improve the applicability of the NeRF methods.

\subsubsection{CLIP}
The conditions from which we want to generate images can exist in various forms, typically in the form of images or texts. To address both cases at the same time, a model that can take multimodal inputs is required. Among such models~\cite{xu2018attngan, tao2020df, clip}, CLIP~\cite{clip} shows an impressive ability to embed text and image information into the same semantic space. We adopt CLIP as our global feature extractor in various conditions, making our model widely applicable for both images and text.



\section{Proposed Approach}


\subsection{Overview}
We propose a novel method called conditional generative NeRF (CG-NeRF), which can generate camera-pose-dependent images conditioned on various types of input data. Unlike recent unconditional generative models that learn neural radiance fields from unlabeled 2D images, we extend the generative model to a conditional model utilizing extra information as input, such as text, sketches, gray-scale, low-resolution images, or even color images.
We design a model that can generate diverse images with different details, sharing the common characteristics of condition inputs.  
As shown in \figref{fig:framework},
the global feature vector $\bf{c}$ extracted from the input condition is fed to the network along with the noise codes $\mathbf{z}^{s}$ and $\mathbf{z}^{a}$ randomly sampled from a standard Gaussian distribution $p_{z}$.
The noise codes specify fine details that are not contained in the given global features.
In the proposed model, the generator $G_{\theta}$ (2 in \figref{fig:framework}) learns radiance field representations and synthesizes images $\hat{\mathbf I}$ corresponding to the given global feature vector $\mathbf{c}$ and noise codes $\mathbf{z}^{s}$ and $\mathbf{z}^{a}$, \textit{i.e.,}
\begin{equation}
G_{\theta}(\boldsymbol{\xi},\mathbf{c}, \mathbf{z}^{s}, \mathbf{z}^{a})=\hat{\mathbf I},
\end{equation}
where $\boldsymbol{\xi}$ is the camera pose for calculating the 3D coordinate $\mathbf{x}$ and the viewing direction $\mathbf{d}$~\cite{GRAF}.
Below, we describe the model structure designed for CG-NeRF in detail.


\begin{table}[]
\centering
\def\arraystretch{1.2}
\captionsetup{width=0.9\linewidth}
\resizebox{0.9\linewidth}{!}{
{\scriptsize
\begin{tabular}{c|l|c}
\hline
 & {\bf Notation} & {\bf Name} \\ \hline\hline
\multirow{5}{*}{\rotatebox[origin=c]{90}{\bf Input}} & $\mathbf{x}\in\mathbb{R}^{3}$ & 3D coordinate \\ \cline{2-3}  & $\mathbf{d}\in\mathbb{R}^{2}$ & Viewing direction \\ \cline{2-3} 
 & $\mathbf{c}\in\mathbb{R}^{L_{c}}$ & Global feautre vector \\ \cline{2-3} 
 & $\mathbf{z}^{s}\in\mathbb{R}^{L_{s}}$ & Shape noise code \\ \cline{2-3} 
 &  $\mathbf{z}^{a}\in\mathbb{R}^{L_{a}}$  & Appearance noise code \\ \hline\hline
\multirow{5}{*}{\rotatebox[origin=c]{90}{\bf Output \quad }} &$\boldsymbol{\gamma}_{i}^{s}, \boldsymbol{\gamma}_{i}^{a}\in\mathbb{R}^{L_{\gamma}}$ & Frequency \\ \cline{2-3} 
 & $\boldsymbol{\beta}_{i}^{s}, \boldsymbol{\beta}_{i}^{a}\in\mathbb{R}^{L_{\beta}}$ & Phase shift \\ \cline{2-3} 
 & $\sigma_{p j}\in\mathbb{R}$  & Density \\ \cline{2-3} 
 & $\mathbf{f}_{p j}\in\mathbb{R}^{L_f}$ & Feature vector \\ \cline{2-3} 
 & $\mathbf{F}_{p}\in\mathbb{R}^{L_f}$ & Rendered feature \\ \cline{2-3} 
 & $\mathbf{I}, \mathbf{\hat{I}}\in\mathbb{R}^{H\times W\times 3}$ & Real/Generated image \\ \hline\hline
\multirow{5}{*}{\rotatebox[origin=c]{90}{\bf Function}} & 
$g_{\theta}$ : $\mathbb{R}^{L_c+L_s+L_a+5} \mapsto {\mathbb{R}^{2L_f}}$
& Feature fields generator \\ \cline{2-3} 
 & $M^s$ : $\mathbb{R}^{L_c+L_s} \mapsto {\mathbb{R}^{N^s\times (L_{\gamma}+L_{\beta})}}$ & Shape mapping network \\ \cline{2-3} 
 &$M^a$ : $\mathbb{R}^{L_c+L_a} \mapsto {\mathbb{R}^{(N^a+1)\times (L_{\gamma}+L_{\beta})}}$& Appearance mapping network \\ \cline{2-3} 
 & $\Phi^s$ : $\mathbb{R}^{3} \mapsto {\mathbb{R}^{L_f}}$ & Shape block \\ \cline{2-3} 
 & $\Phi^a$ : $\mathbb{R}^{L_f+2} \mapsto {\mathbb{R}^{2L_f}}$ & Appearance block \\ \hline
\end{tabular}
}
}
\caption{\small Summarized notations. $p\in\{1, \,\cdots\, , H_{V}W_{V}\}$, and $j\in\{1,\,\cdots\, , J\}$. $J$ indicates the number of sampling points per ray. $H \times W$ and $H_V \times W_V$ are the spatial resoultion of image and features, respectively. }
\label{tb:notation}
\end{table}

\subsection{Architecture: Conditional Generative NeRF }

As illustrated in \Figref{fig:framework}, the main architecture consists of three components: a (1) feature extractor $E$ that extracts global feature vectors from the given conditions, (2) a generator that creates an image by reflecting the conditions, and (3) a discriminator that distinguishes real images from fake images based on the condition input and that predicts the camera poses of fake images for the PD loss, which will be described in detail later. 




As CG-NeRF aims to synthesize conditional 3D-aware images, the condition input is encoded to a global feature vector through the global feature extractor $E$ (1 in \figref{fig:framework}). To extract global semantic features from the given condition inputs in our case, we adopt CLIP~\cite{radford2021learning}, accomodating various types of input conditions such as images and text, as a state-of-the-art multimodal encoder. 




We design our generator network by combining two recent promising techniques, which are proven to generate high-quality images for the generative neural radiance field task: a SIREN-based backbone ~\cite{pigan} and a feature-level volume rendering method ~\cite{GIRAFFE}. 
The SIREN-based ~\cite{sitzmann2020implicit} network architecture enhances the visual quality of the NeRF-based generative model but requires a large amount of memory for training due to color-level volume rendering at the full image resolution ~\cite{pigan}. To address this issue, we leverage feature-level volume rendering, inspired by a recently proposed method ~\cite{GIRAFFE}. The feature-level volume rendering process substantially mitigates the problem because a volume is rendered at the level of feature vector $\mathbf{f}$, having a smaller scale than the image resolution. 




Given a global feature vector $\mathbf{c}$, a noise code of shape $\mathbf{z}^{s}$ and appearance $\mathbf{z}^{a}$, the feature fields generator $g_{\theta}$ (2-1 in~\figref{fig:framework}) produces the density $\sigma$ and feature vector $\mathbf{f}$ in the corresponding $\mathbf{x}$ and $\mathbf{d}$ as  
\begin{equation}
g_{\theta}({\bf x}_{p j},{\bf d}_{p}, \mathbf{c}, \mathbf{z}^{s}, \mathbf{z}^{a})=(\sigma_{p j},{\bf f}_{p j}),
\end{equation}
where $\sigma_{p j}$ and ${\bf f}_{p j}$ denote the density and the feature vector, respectively, at the corresponding 3D coordinate. Further details are described in the next section. 


Once the density $\sigma$ and the feature vector $\bf{f}$ are estimated by the feature fields generator $g_{\theta}$ (2-1 in ~\figref{fig:framework}) at each 3D coordinate, the final feature $\bf{F}_{p} \in\mathbb{R}^{L_{f}}$ is computed through a feature-level volume rendering process as
\begin{equation}
\mathbf{F}_{p}=\sum_{j=1}^{J} T_{p j} \alpha_{p j} \mathbf{f}_{p j}, \\
\end{equation}
where the transmittance $T_{p j}=\prod_{k=1}^{j-1}\left(1-\alpha_{p k}\right)$. The alpha value for $\mathbf{x}_{p j}$ is calculated as $\alpha_{p j}=1-e^{-\sigma_{p j} \delta_{p j}}$, and $\delta_{p j}$ is the distance between neighboring sample points along the ray direction~\cite{mildenhall2020nerf}. 
The 2D feature map $\mathbf{F}\in\mathbb{R}^{H_{V}\times W_{V}\times L_{f}}$ rendered through the volume rendering process is then upsampled to a RGB images at a higher resolution $\hat{\mathbf{I}}\in\mathbb{R}^{H\times W\times 3}$ using the 2D convolutional neural network (CNN) decoder network (2-2 in ~\figref{fig:framework}).
The decoder network consists of CNN layers with leaky ReLU activation functions ~\cite{xu2015empirical} and nearest neighbor upsampling layers. 





\subsection{Condition-based Disentangling Network}


We propose a novel approach that aims to disentangle both the shape and appearance contained in a given global feature vector. For a text condition example `` round bird with a red body", ``round" and ``bird" are shapes, and ``red" is an attribute indicating the appearance. 
Two mapping networks $M^s$ and $M^a$ serve to generate the styles of the shape and appearance, respectively, from the global feature vector $\mathbf{c}$ and noise codes $\mathbf{z}^{s}$ and $\mathbf{z}^{a}$. The global feature vector $\mathbf{c}\in\mathbb{R}^{L_c}$ contains the prominent attribute of the condition. In constrast, the noise codes $\mathbf{z}^s\in\mathbb{R}^{L_s}$ and $\mathbf{z}^a\in\mathbb{R}^{L_a}$ are responsible for the details that the global feature vector does not include. The mapping network consists of pairs of a linear layer and ReLU and produces frequencies $\boldsymbol{\gamma}$ and phase shifts $\boldsymbol{\beta}$ as
\begin{equation}
\begin{aligned}
&M^{s}\left(\mathbf{c}, \mathbf{z^{s}}\right)=\text{cat}\{\left(\boldsymbol{\gamma}_{i}^{s}, \boldsymbol{\beta}_{i}^{s}\right)\}_{i=1 \cdots N^{s}}\\
&M^{a}\left(\mathbf{c}, \mathbf{z^{a}}\right)=\text{cat}\{\left(\boldsymbol{\gamma}_{i}^{a}, \boldsymbol{\beta}_{i}^{a}\right)\}_{i=1 \ldots N^{a}+1} \text {, }
\end{aligned}
\end{equation}
where $N^s$ and $N^a$ denote the numbers of MLPs in each block. 
\emph{cat} indicates channel-wise concatenation.
The predicted frequencies and phase shifts are fed to the two blocks $\Phi^s$ and $\Phi^a$ in the feature fields generator. 



Taking these as inputs along with the 3D coordinate $\mathbf{x}$ and the direction $\mathbf{d}$, two consecutive blocks encode features using pairs of a linear layer and activation function of feature-wise linear modulation (FiLM) SIREN. The sine function of the FiLM SIREN layer modulated by the obtained frequency and phase shift are applied to the outputs of the linear layers as an activation function; \textit{i.e.,} 
\begin{equation}
\begin{aligned}
\phi_{i}\left(\mathbf{y}_{i}\right)=\sin (\boldsymbol{\gamma}_{i}(\mathbf{W}_{i} \mathbf{y}_{i}+\mathbf{b}_{i})+\boldsymbol{\beta}_{i}) \text{,}
\end{aligned}
\end{equation}
where $\phi_{i}$ : $\mathbb{R}^{M_i} \mapsto {\mathbb{R}^{N_i}}$ is the $i$-th MLP of each ${\Phi}^s$ and ${\Phi}^a$. $\mathbf{W}_i\in\mathbb{R}^{N_i \times {M_i}}$ and $\mathbf{b}_i\in\mathbb{R}^{N_i}$ are the weight and the bias 
applied to input $\mathbf{y}_{i}\in\mathbb{R}^{M_i}$. 
The two blocks in the feature fields generator have the following formulations:
\begin{equation}
\begin{aligned}
&\resizebox{.55\hsize}{!}{${\Phi}^{s}\left(\mathbf{x}_{p j}\right)=\phi_{N^{s}}^{s}\left(\phi_{N^{s}-1}^{s}(\cdots \phi_{1}^{s}(\mathbf{x}_{p j}))\right)\text{,}$}\\
&\resizebox{.8\hsize}{!}{${\Phi}^{a}({\Phi}^{s}(\mathbf{x}_{p j}),\mathbf{d}_{p})=\phi_{N^{a}+1}^{a}(\text{cat}(\phi_{N^{a}}^{a}(\cdots \phi_{1}^{a}({\Phi}^{s}\left(\mathbf{x}_{p j}\right))),\mathbf{d}_{p}))\text{.}$}\\
\end{aligned}
\end{equation}

Inspired by an existing approach~\cite{GRAF}, we assign the roles of reflecting the shape to the first block, close to the input, and the appearance to the second block, close to the output. The block for shape utilizes the 3D coordinate as the input to generate shape-encoded features, while the appearance block takes the output of the previous block as input and generates encoded features of the shape and appearance. By utilizing these features and viewing directions as inputs, features reflecting the viewing direction are generated from the last layer of the appearance block.

\subsection{Pose-consistent Diversity Loss} 

As our method generates images conditioned on extra inputs, variations of the output images are restricted, especially when a color image is given as a condition input. To enable the generator network to produce semantically diverse images based on the condition input, we regularize the generator network with the diversity-sensitive loss ~\cite{yang2019diversity}. This is defined as
\begin{equation}
\begin{aligned}
\mathcal{L}_{\text{div}}(\theta)=\mathbb{E}_{ \mathbf{z}^{s},\mathbf{z}^{a} \sim p_{z},\boldsymbol{\xi} \sim p_{\xi},\mathbf{c}\sim p_{r}}[\parallel \hat{\mathbf{I}}_{1} - \hat{\mathbf{I}}_{2} \parallel_{1}],
\end{aligned}
\end{equation}
where $\hat{\mathbf{I}}_{1}$ is $G_{\theta}(\boldsymbol{\xi},\mathbf{c}, \mathbf{z}^{s1},\mathbf{z}^{a1})$ and $\hat{\mathbf{I}}_{2}$ is $G_{\theta}(\boldsymbol{\xi},\mathbf{c}, \mathbf{z}^{s2},\mathbf{z}^{a2})$.

However, we empirically discover that simply applying the diversity-sensitive loss causes undesirable effects that attempt to change not only the style but also the pose of the output images (\figref{fig:abl-pose}). Because the pose of the output images should be determined only by the input camera pose $\boldsymbol{\xi}$,
pose changes in the output images are a significant side effect. We analyze this undesirable phenomenon as follows; from the generator network's point of view, the model maximizes the pixel difference via two different methods: (1) changing the style of the output images as desired or (2) changing the poses between two output images generated with the same camera pose, which is strongly undesired. 




To explicitly address such an issue, we propose a pose regularization term applicable to the original diversity-sensitive loss, which explicitly penalizes pose difference between images generated from different noise codes $\mathbf{z}^{s}$ and $\mathbf{z}^{a}$ but from the same camera pose. The intuition behind the proposed regularization is that the model generates two images to have only a style difference constrained to have the same pose, which can be additionally learned by an auxiliary network.
We propose to add the regularization term $\mathcal{L}_{\text{pose}}$ to the diversity-sensitive loss $\mathcal{L}_{\text{div}}$, which is defined as 
\begin{equation}
\resizebox{.8\hsize}{!}{$\mathcal{L}_{\text{pose}}(\theta)=\mathbb{E}_{\mathbf{z}^{s},\mathbf{z}^{a} \sim p_{z},\boldsymbol{\xi} \sim p_{\xi},\mathbf{c}\sim p_{r}}[1- \cos(D_{\psi}^{\xi}(\hat{\mathbf{I}}_{1})-D_{\psi}^{\xi}(\hat{\mathbf{I}}_{2}))],$}
\end{equation}

where $D_{\psi}^{\xi}$ is the auxiliary pose estimator network we additionally train for the pose penalty loss jointly with the discriminator. 

The proposed method simultaneously learns the output images' poses by training the pose estimator network. We modify our discriminator network to contain an auxiliary pose estimator, by adjusting the channel size of the last layer to estimate the camera pose values of the output image. Because we randomly sample camera poses $\boldsymbol{\xi}$ from the prior distribution $p_{\xi}$ to generate view-consistent images, the sampled camera pose is utilized as the ground truth pose when training the pose estimator. We define the camera pose $\boldsymbol{\xi}$ with radius $r_{cam}$, rotation angle $\kappa_{r} \in [-\pi,\pi]$, and elevation angle $\kappa_{e} \in [0,\pi]$. Given that we use a fixed value for $r_{cam}$=1, the pose estimator predicts the rotation angle and elevation angle, applying the Sigmoid function to the output value multiplied by $2\pi$ and $\pi$ respectively. 
The camera pose reconstruction loss is defined as
\begin{equation}
\begin{aligned}
\resizebox{.8\hsize}{!}{$\mathcal{L}_{\text{pose}}(\psi)=\mathbb{E}_{\mathbf{z}^{s},\mathbf{z}^{a} \sim p_{z},\boldsymbol{\xi} \sim p_{\xi},\mathbf{c}\sim p_{r}}[ 1- \cos(D_{\psi}^{\xi}(\hat{\mathbf{I}})-\boldsymbol{\xi}_{\text{gt}})],$}\\
\end{aligned}
\end{equation}

where $D_{\psi}^{\xi}(\hat{\mathbf{I}})=\boldsymbol{\xi}_{\text{pred}} = (\hat{\kappa_{r}}, \hat{\kappa_{e}})$. $D_{\psi}^{\xi}$ denotes the auxiliary pose estimator and $\boldsymbol{\xi}_{\text{gt}}$ is a randomly sampled camera pose value to generate $\hat{\mathbf{I}}$.
Because the angle can be represented by a periodic function, 
we design the pose reconstruction loss with the cosine function to penalize the angle difference, addressing its discontinuity at $2\pi$. 

\subsection{Training Objective}

To synthesize conditional outputs, we adopt a conditional GAN~\cite{isola2017image} by training a discriminator that learns to match images and condition feature vectors. As shown in \Figref{fig:framework}, the discriminator extracts the image feature through a series of 2D convolution layers, and the image feature is then concatenated with matching condition $\mathbf{e}$ to predict the condition-image semantic consistency.
The matching condition $\mathbf{e}\in\mathbb{R}^{L_c+L_s+L_a}$ is the global feature vector $\mathbf{c}$ concatenated with detail codes $\mathbf{z^{s}}$ and $\mathbf{z^{a}}$.
The number of feature extracting layers is determined by the resolution of the training images. The discriminator network learns whether the given image is real or fake and matches its condition feature vector simultaneously.

At training time, we use the non-saturating GAN loss with a matching-aware gradient penalty~\cite{mescheder2018training, tao2020df}. Instead of the $R_{1}$ gradient penalty~\cite{mescheder2018training}, we adopt the matching-aware gradient penalty loss, which is known to promote the generator to synthesize more realistic and semantic-consistent images to condition-image pairs. We define three different types of data items: synthetic images with the matching condition, real images with a matching condition, and real images with a mismatching condition. The target data point on which the gradient penalty is applied can be defined by real images with the matching condition feature vector. The entire formulation of conditional GAN loss, \textit{i.e.,}
\begin{equation}
\begin{aligned}
&\mathcal{L}_{\text{adv}}(\psi) =\mathbb{E}_{\mathbf{I} \sim p_{r}}\left[f\left(D_{\psi}(\mathbf{I}, \mathbf{e})\right)\right]\\
&+(1 / 2) \mathbb{E}_{\mathbf{I} \sim p_{mis}}\left[f\left(-D_{\psi}(\mathbf{I}, \mathbf{e})\right)\right]\\
&+(1 / 2) \mathbb{E}_{\boldsymbol{\xi} \sim p_{\xi},\mathbf{e}\sim p_{r},p_{z} }\left[f\left(-D_{\psi}\left(G_{\theta}(\boldsymbol{\xi},\mathbf{c}, \mathbf{z}^{s},\mathbf{z}^{a}), \mathbf{e}\right)\right)\right] \\
&+k \mathbb{E}_{\mathbf{I} \sim p_{r}}\left[\left(\left\|\nabla_{\mathbf{I}} D_{\psi}(\mathbf{I}, \mathbf{e})\right\|+\left\|\nabla_{\mathbf{e}} D_{\psi}(\mathbf{I}, \mathbf{e})\right\|\right)^{p}\right], \\
&\mathcal{L}_{\text{adv}}(\theta) =\mathbb{E}_{\boldsymbol{\xi} \sim p_{\xi},\mathbf{e}\sim p_{r},p_{z} }\left[f\left(D_{\psi}\left(G_{\theta}(\boldsymbol{\xi},\mathbf{c}, \mathbf{z}^{s},\mathbf{z}^{a}), \mathbf{e}\right)\right)\right] \\
\end{aligned}
\label{eqn:eqn-adv}
\end{equation}

where $f(u)=-\log (1+\exp (-u))$. $p_{r}$ and  $p_{mis}$ denote the real data distribution and mismatching data distribution, respectively. $k$ and $p$ are two hyper-parameters that balance the gradient penalty effects. 



Our full training objective functions for the generator network $G_{\theta}$ are summarized as 
\begin{equation}
\begin{aligned}
&\mathcal{L}_{\text{total}}=\mathcal{L}_{\text{adv}}-\lambda_{\text{div}}\mathcal{L}_{\text{div}}+\lambda_{\text{pose}}\mathcal{L}_{\text{pose}},\\
\end{aligned}
\end{equation}

where $\lambda_{\text{div}}$ and $\lambda_{\text{pose}}$ are weights for each loss term. 



\begin{figure}[t]
\captionsetup{width=0.9\linewidth}
\def\arraystretch{0.2}
\begin{center}
\begin{tabular}{@{}c@{}c@{}c}
\includegraphics[width=0.2\linewidth]{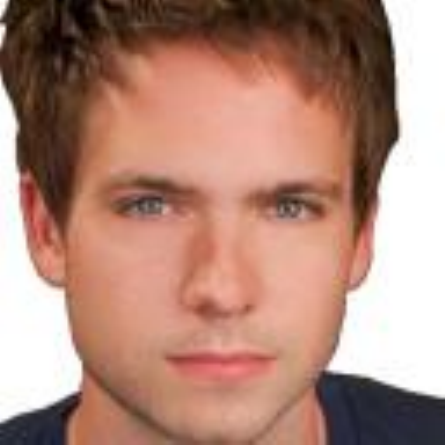} &
\includegraphics[width=0.2\linewidth]{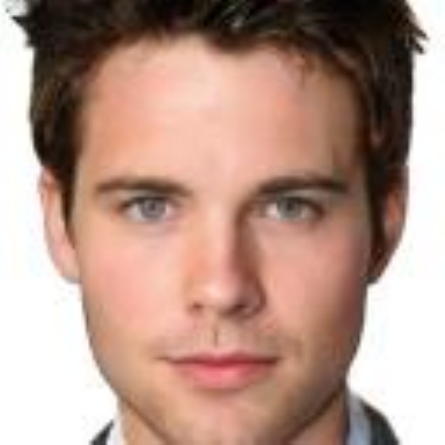} &
\includegraphics[width=0.5\linewidth]{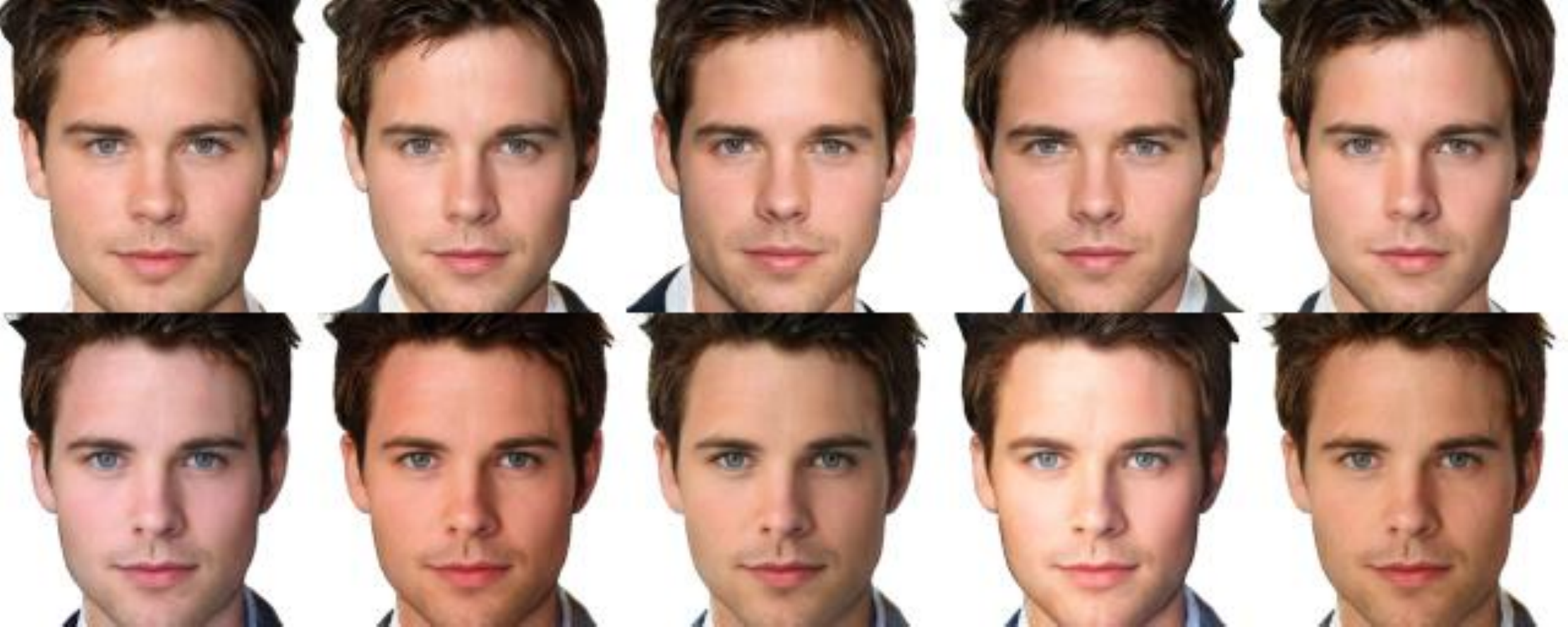} \\

\includegraphics[width=0.2\linewidth]{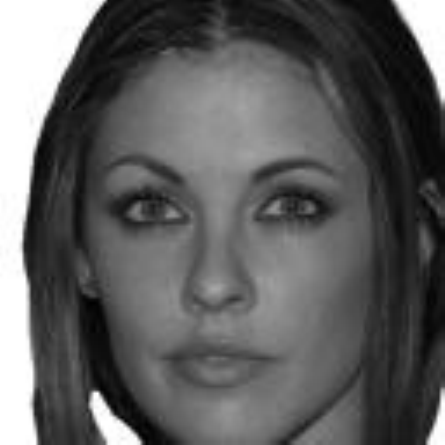} &
\includegraphics[width=0.2\linewidth]{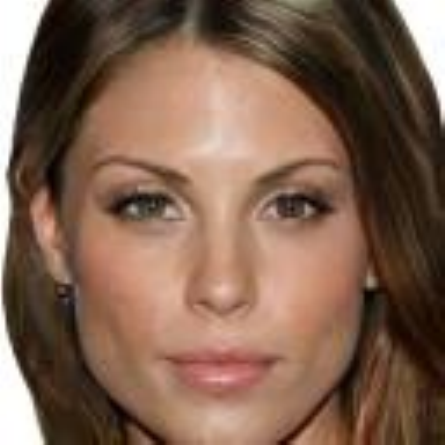} &
\includegraphics[width=0.5\linewidth]{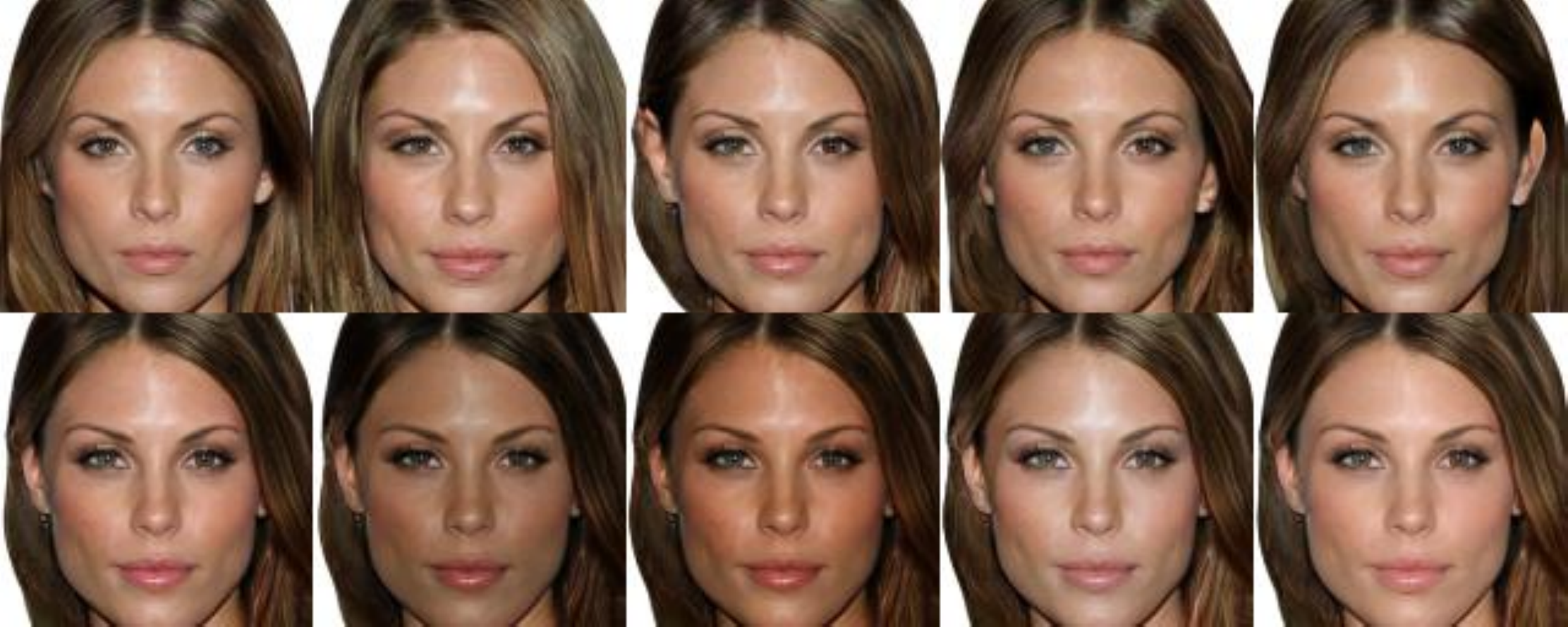} \\

\includegraphics[width=0.2\linewidth]{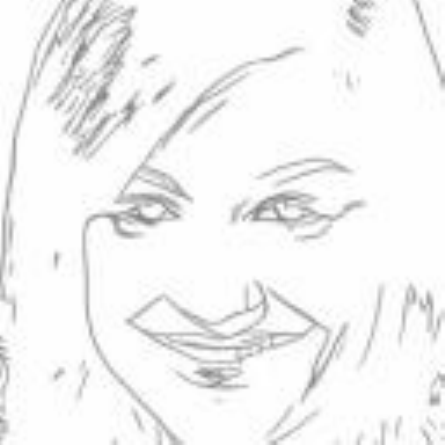} &
\includegraphics[width=0.2\linewidth]{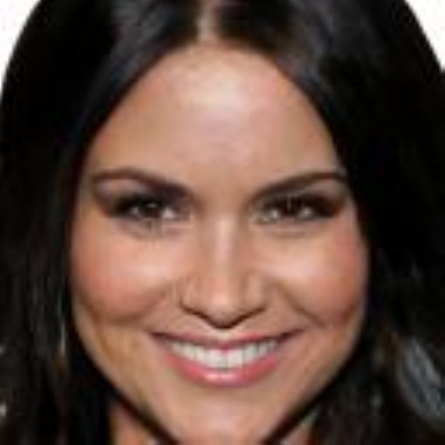} &
\includegraphics[width=0.5\linewidth]{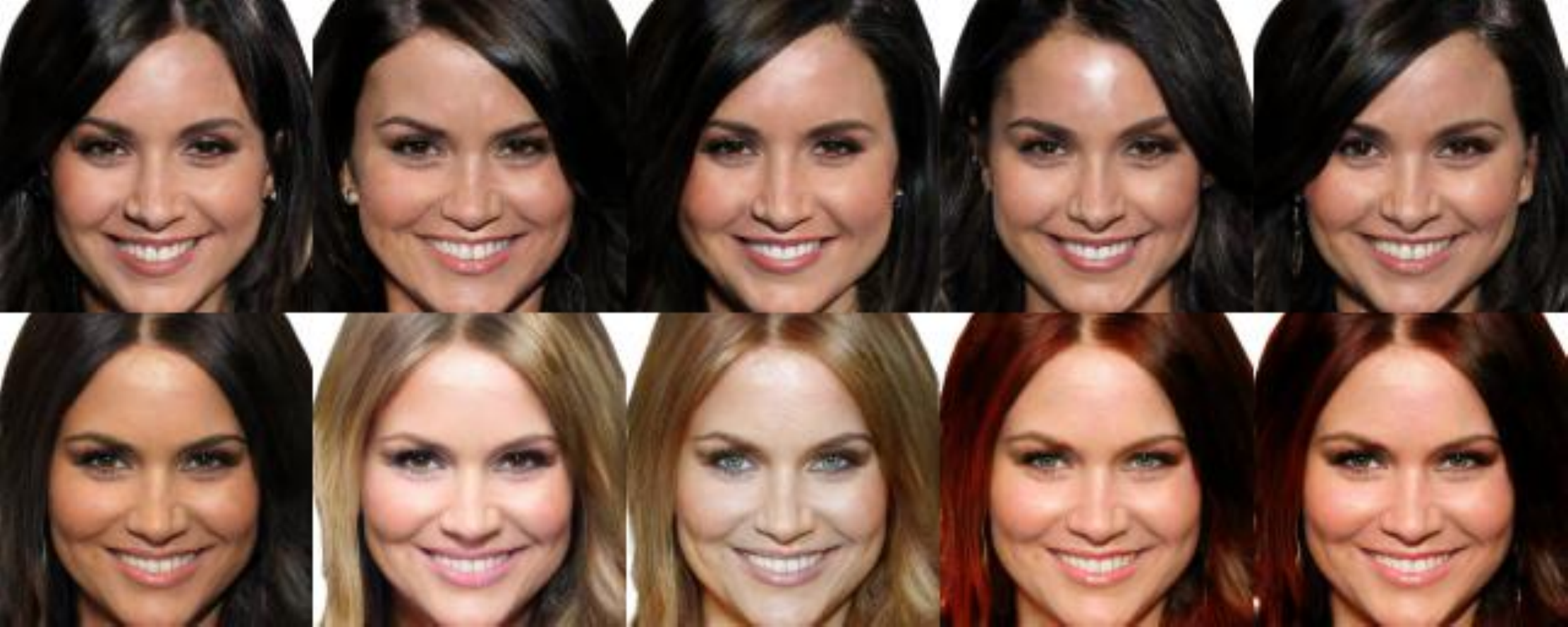} \\

\includegraphics[width=0.2\linewidth]{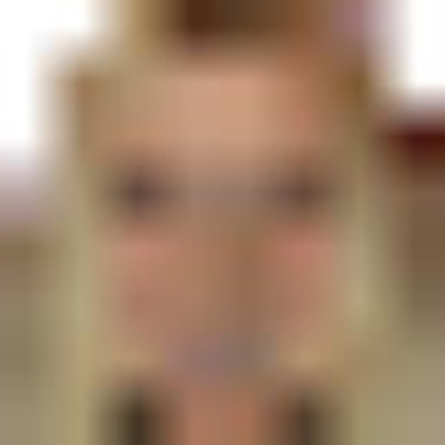} &
\includegraphics[width=0.2\linewidth]{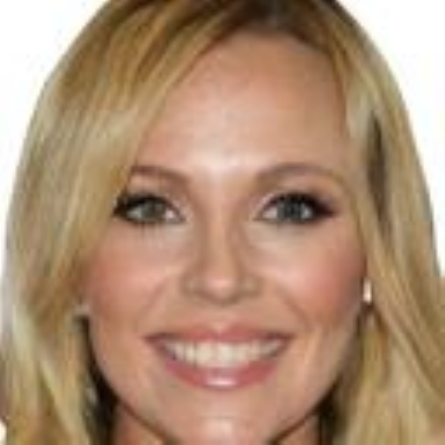} &
\includegraphics[width=0.5\linewidth]{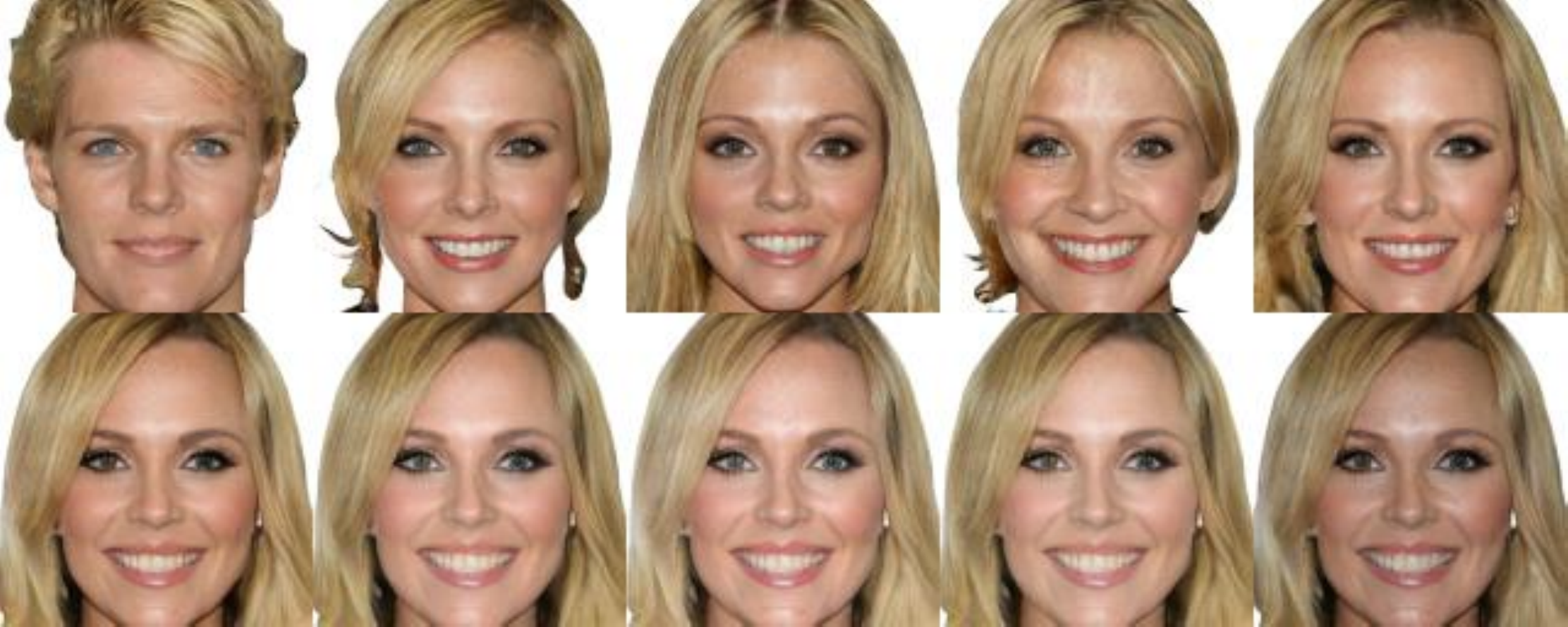} \\

\includegraphics[width=0.2\linewidth]{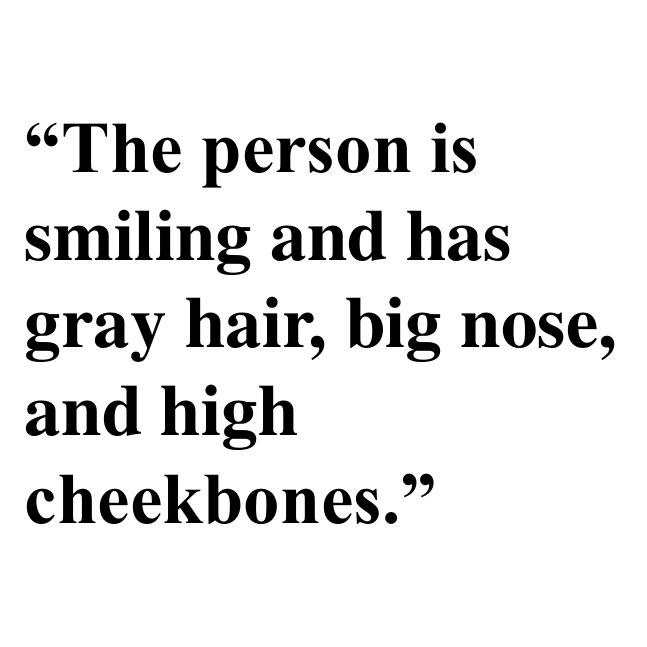} &
\includegraphics[width=0.2\linewidth]{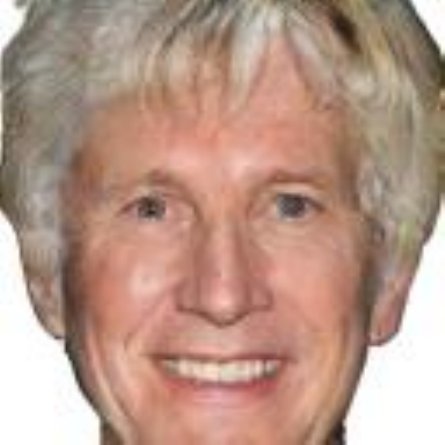} &
\includegraphics[width=0.5\linewidth]{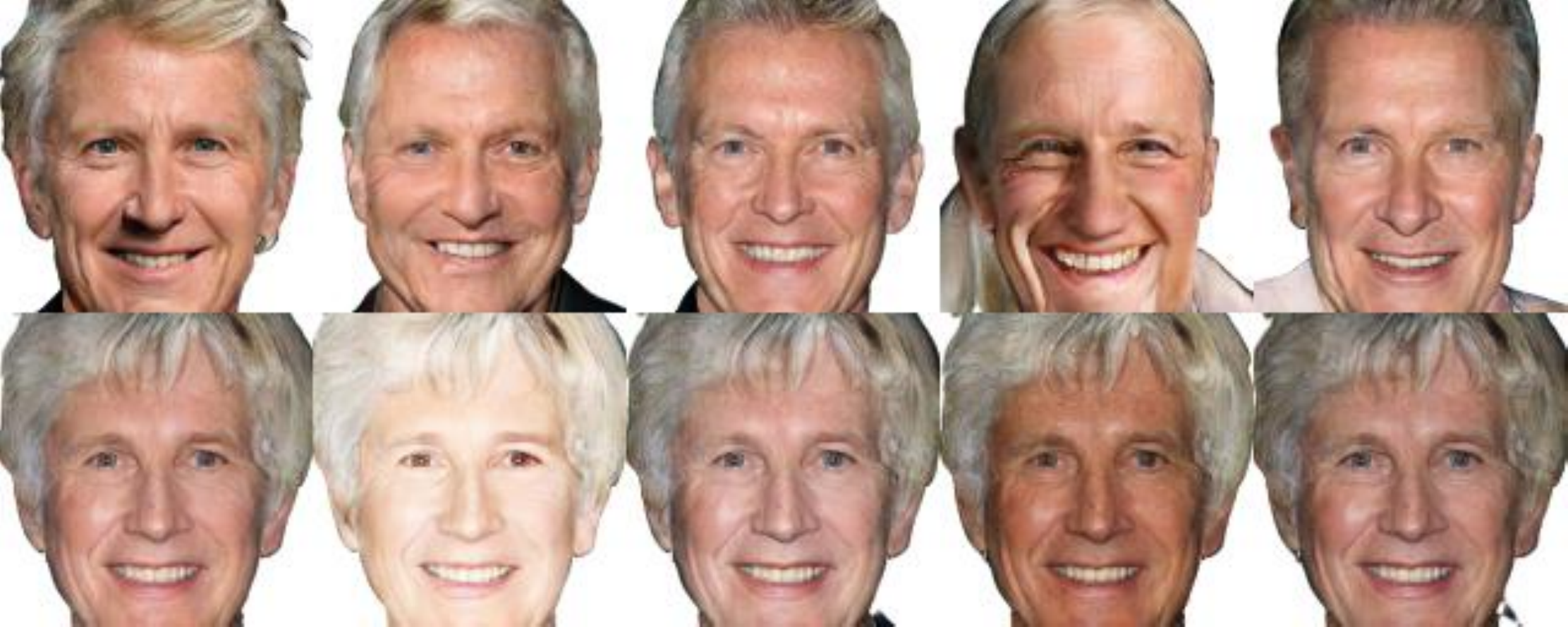} \\

{\small Input}&{\small Average}&{\small Random Shape / Appearance}
\end{tabular}
\end{center}
\caption{\small Qualitative results with various condition input. For each condition type, the average output image generated with zero-value noise codes and output images generated from five different shape noise codes (in row 1) and appearance noise codes (in row 2) are visualized.}
\label{fig:quali-condition}
\end{figure}
\section{Experiments}


\subsubsection{Dataset setups}
We evaluate our CG-NeRF on various datasets, in this case CelebA-HQ~\cite{karras2017progressive}, CUB-200~\cite{wah2011caltech}, and Cats ~\cite{zhang2008cat}. Bacause the proposed method generates output images with extra input information given, unlike NeRF-based generative models~\cite{pigan,GIRAFFE, GRAF}, we evaluate our method only on the test set for a fair comparison.  
For the condition inputs, we select five different data forms to consider the different properties of input conditions in terms of the shape and appearance, \textit{e.g.,} color images, grayscale, sketches, low-resolution images, and text. To generate 3D-aware images from sketch conditions, first we apply a Sobel filter to extract pseudo sketch information from the image~\cite{richardson2021encoding} after which we apply a sketch simplification method ~\cite{simo2016learning}. For low-resolution image conditions, we apply bilinear downsampling to images with a ratio of 1/16. 
Training images are resized to a resoultion of 128$\times$128. 
To extract the global feature only of the object, 
we remove the background for CelebA-HQ and CUB-200 datasets. 

\subsection{Experimental results}

\begin{table*}[t]
\centering
\resizebox{0.95\linewidth}{!}{\huge
\centering
\begin{tabular}{c|l|c|c|c|c||c|c|c|c||c|c|c|c||c|c|c|c}
\hline
\multicolumn{2}{c||}{\multirow{2}{*}{Dataset}} & \multicolumn{4}{c||}{CelebA} & \multicolumn{4}{c||}{CelebA-HQ} & \multicolumn{4}{c||}{Cats} & \multicolumn{4}{c}{CUB-200} \\ \cline{3-18} 
\multicolumn{2}{c||}{} & \multicolumn{4}{c||}{with background} & \multicolumn{4}{c||}{with background} & \multicolumn{4}{c||}{with background} & \multicolumn{4}{c}{without background} \\ \hline
\multicolumn{2}{c||}{Method} & \begin{tabular}[c]{@{}c@{}}Image \\ Resolution\end{tabular} & FID↓ & Precision↑ & Recall↑ & \begin{tabular}[c]{@{}c@{}}Image \\ Resolution\end{tabular} & FID↓ & Precision↑ & Recall↑ & \begin{tabular}[c]{@{}c@{}}Image \\ Resolution\end{tabular} & FID↓ & Precision↑ & Recall↑ & \begin{tabular}[c]{@{}c@{}}Image \\ Resolution\end{tabular} & FID↓ & Precision↑ & Recall↑ \\ \hline
\multicolumn{2}{c||}{GRAF} & 128 & 53.13 & 0.75 & 0.01 & 256 & 64.35 & 0.57 & 0.00 & 64 & {\bf 13.73} & 0.86 & 0.20 & 64 & 41.65 & 0.80 & 0.09 \\ \hline
\multicolumn{2}{c||}{GIRAFFE} & 64 & 25.39 & \textbf{0.88} & 0.10 & 256 & 38.16 & 0.69 & 0.03 & 64 & 16.05 & 0.74 & 0.37 & - & - & - & - \\ \hline
\multicolumn{2}{c||}{pi-GAN} & 128 & \textbf{20.92} & 0.75 & \textbf{0.49} & - & - & - & - & 128 & 22.57 & 0.61 & 0.25 & - & - & - & - \\ \hline
\multicolumn{2}{c||}{-} & \multicolumn{4}{c||}{-} & \multicolumn{4}{c||}{\begin{tabular}[c]{@{}c@{}}without background,\\  conditioned on color image\end{tabular}} & \multicolumn{4}{c||}{\begin{tabular}[c]{@{}c@{}}with background,\\  conditioned on color image\end{tabular}} & \multicolumn{4}{c}{\begin{tabular}[c]{@{}c@{}}without background,\\  conditioned on text\end{tabular}} \\ \hline

\multicolumn{2}{c||}{\multirow{2}{*}{\textbf{Ours}}} & \multirow{2}{*}{-} & \multirow{2}{*}{-} & \multirow{2}{*}{-} & \multirow{2}{*}{-} & \multirow{2}{*}{128} & \multirow{2}{*}{\textbf{7.01}} & \multirow{2}{*}{\textbf{0.90}} & \multirow{2}{*}{\textbf{0.55}} & \multirow{2}{*}{128} & \multirow{2}{*}{\textbf{13.86}} & \multirow{2}{*}{\textbf{0.91}} & \multirow{2}{*}{\textbf{0.52}} & \multirow{2}{*}{128} & \multirow{2}{*}{\textbf{26.53}} & \multirow{2}{*}{\textbf{0.82}} & \multirow{2}{*}{\textbf{0.22}} \\ [2ex]\hline
\end{tabular}
}
\caption{\small Quantitative comparison in terms of FID, precision, and recall. A low FID score means that the distribution of the generated image is close to that of the real image in terms of the mean and standard deviation. A high precision score implies that the generated image is realistic, and a high recall score indicates that the generated images capture greater variation of the real images. To guarantee the optimal performance of each algorithm, we measure the performance based on publicly available pre-trained models from previous work. 
}
\label{tb:quantative_prev}
\end{table*}

\begin{figure}[t]
\captionsetup{width=0.9\linewidth}
\centering
\def\arraystretch{0.2}
\begin{tabular}{@{}c}
\includegraphics[width=0.9\linewidth]{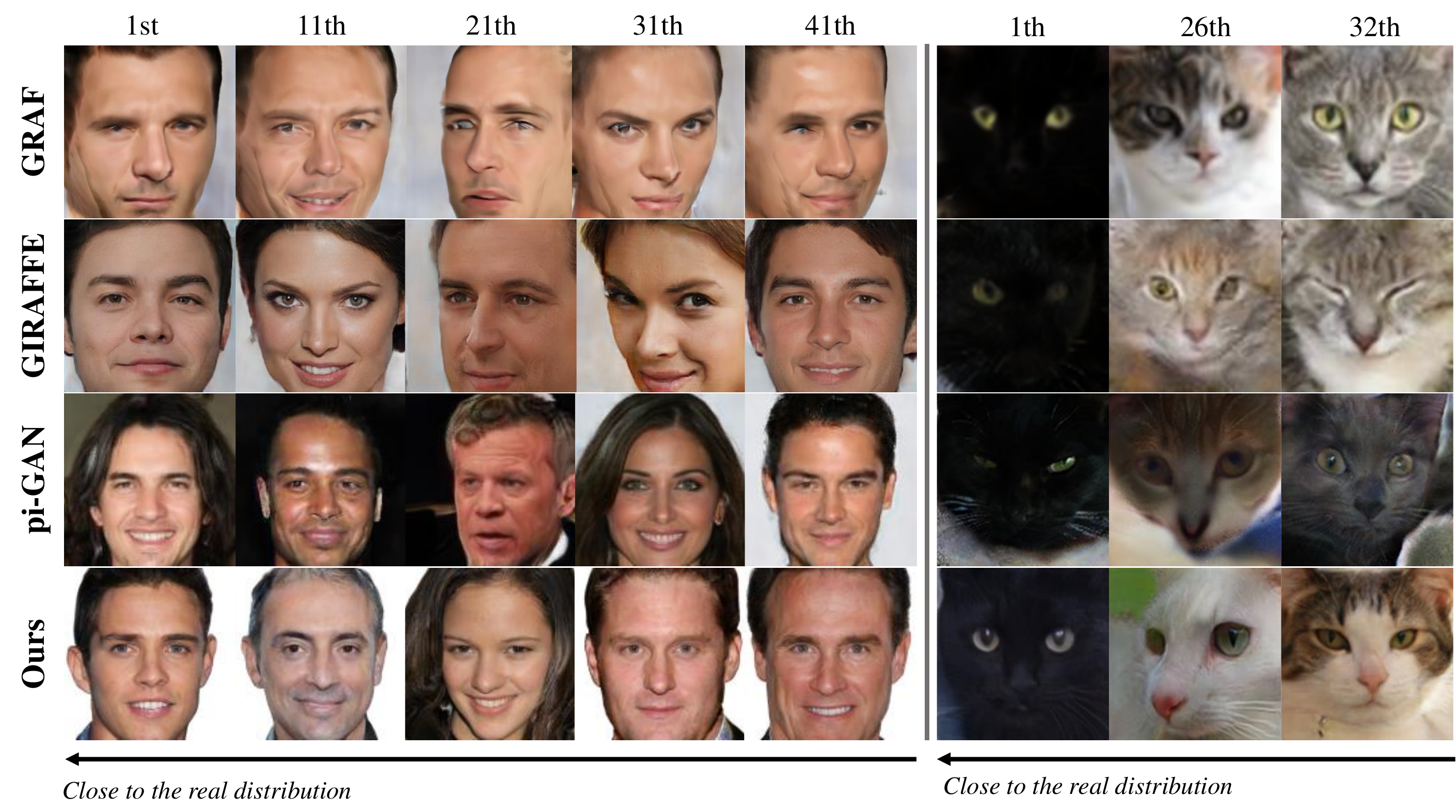} \\
\end{tabular}

\caption{\small Comparison of qualitative results to previous studies on the CelebA (pi-GAN), CelebA-HQ (GRAF, GIRAFFE, Ours), and Cats. For each dataset, the distance between the generated and the real image increases from left to right. For the Cats dataset, since black cat images are mainly occupied at the top ranks, we sample results with the larger rank interval than the CelebA (CelebA-HQ) dataset.
}
\label{fig:quali_prev}
\end{figure}

To the best of our knowledge, there exists a no previous work performing condition generative NeRF task has been published.  Hence, we perform quantitative and qualitative comparison of our model with existing NeRF-based generative models ~\cite{GRAF,GIRAFFE,pigan} to demonstrate the competitive performance of the proposed method. 

\subsubsection{Quantitative comparison}
To evaluate our approach quantitatively, we measure three metrics: the Frechet Inception Distance (FID) ~\cite{heusel2017gans}, precision, and recall using publicly available libraries\footnote{https://github.com/toshas/torch-fidelity}\footnote{https://github.com/clovaai/generative-evaluation-prdc} ~\cite{obukhov2020torchfidelity,ferjad2020icml}.
FID is the most popular metric for evaluating the quality of GANs as it reveals a discrepancy between distributions of real and fake images. On the other hand, precision and recall measure the quality of GANs in terms of fidelity and diversity, respectively. 
As reported in~\tabref{tb:quantative_prev}, to guarantee the most reliable performance of the previous methods, we evaluate the comparison results using a publicly available pre-trained model and its corresponding experiment setting.
Because these visual quality metrics are highly sensitive to measurement tools showing incorrect implementations across different image processing libraries~\cite{parmar2021buggy}, we evaluate all of the comparison methods using the same measurement tools. 
Based on the performances we measured, the proposed method shows better scores in terms of FID, precision, and recall compared to the existing methods for the most part.
For the Cats dataset, which has all four comparison results, our method still produces competitive performance on FID as well as the best performance on precision and recall.

\subsubsection{Qualitative comparison}
\Figref{fig:quali_prev} shows comparisons of our method with other NeRF-based generative models in terms of the visual quality. For a fair comparison, according to the definition of precision ~\cite{kynkaanniemi2019improved}, we select images in the order of the closest distance to the real image among fake images existing in the manifold of the real image. The distance is measured utilizing features of the real and fake images in the Euclidean space due to the high dimensionality of the image and lack of semantics in the RGB space. To display diverse images across all the methods, images are sampled with the distanced rank interval. 
Our method shows competitive visual quality on both the CelebA (CelebA-HQ) and Cats datasets. (\figref{fig:quali_prev}).

\begin{table}[!t]
\centering
\captionsetup{width=0.9\linewidth}
\resizebox{0.45\linewidth}{!}{
\begin{tabular}{l||c|c}
\hline

\textbf{CelebA-HQ}  & FID↓ & IS↑\\ \hline

Color Image & 7.01 & 2.14  \\ \hline

Grayscale & 7.23 & 2.12 \\ \hline

Sketch  & 7.01 & 2.16  \\ \hline

Low-Resolution & 7.91 & 2.05 \\ \hline

Text  & 7.31 & 2.13  \\ \hline

\end{tabular}
}
\resizebox{0.45\linewidth}{!}{
\begin{tabular}{l||c|c}
\hline

\textbf{Cats}    & FID↓ & IS↑  \\ \hline

Color Image & 13.86 & 2.06  \\ \hline

Grayscale & 12.51 & 2.02 \\ \hline


Low-Resolution & 19.40 & 2.13 \\ \hline\hline


\textbf{CUB-200} & FID↓ & IS↑ \\ \hline


Text & 26.53 & 3.52  \\ \hline

\end{tabular}
}
\caption{\small Quantitative comparisons (FID / IS) on the CelebA-HQ, Cats, and CUB-200 datasets with different condition types in terms of the image quality.}
\label{tb:quanti-condition}
\end{table}



\subsubsection{Effects of various condition types}


In this section, we perform experiments to analyze the training behavior of our method depending on the input condition type. 
We compare the results with five different types of condition input to validate that our method yields consistent generation performance. 
As shown in~\figref{fig:quali-condition}, as the color image has the largest amount of condition information among the five different condition types, it restricts the range of style variation of output images generated with random noise codes. In contrast, weak conditions such as text or low-resolution images show dynamic changes in their results with random shapes or appearances. 
To evaluate our approach in terms of condition types quantitatively, we measure the FID ~\cite{heusel2017gans} and Inception Score (IS) ~\cite{salimans2016improved} as shown in \tabref{tb:quanti-condition}. For each dataset, our method consistently maintains high visual quality across all types of input conditions. 

\subsection{Analysis of Experiments}


\begin{figure}[t]
\captionsetup{width=0.9\linewidth}
\centering
\resizebox{0.84\linewidth}{!}{
\begin{tabular}{@{}c@{\hskip 0.02\linewidth}c}
\includegraphics[width=0.45\linewidth]{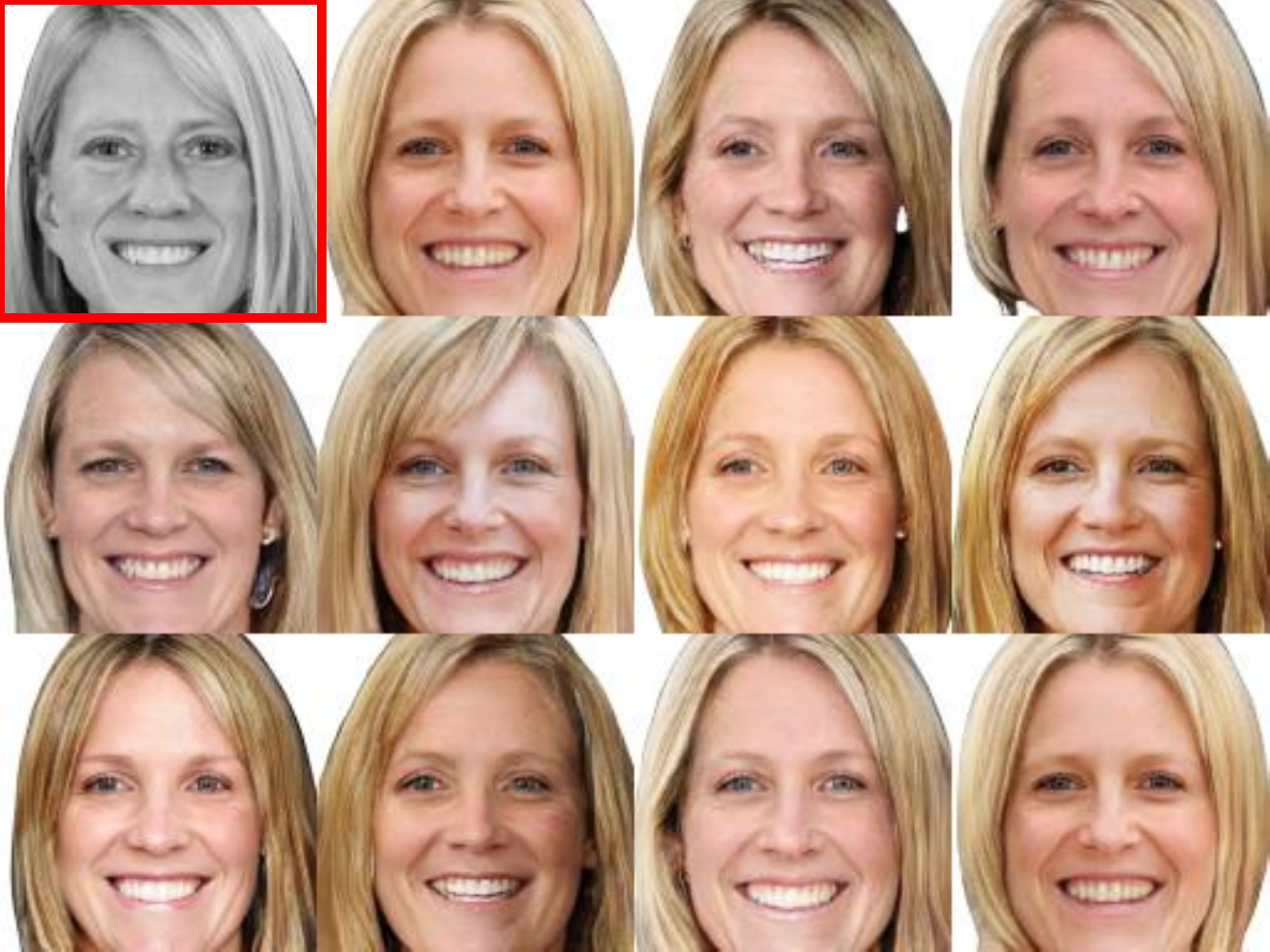} &
\includegraphics[width=0.45\linewidth]{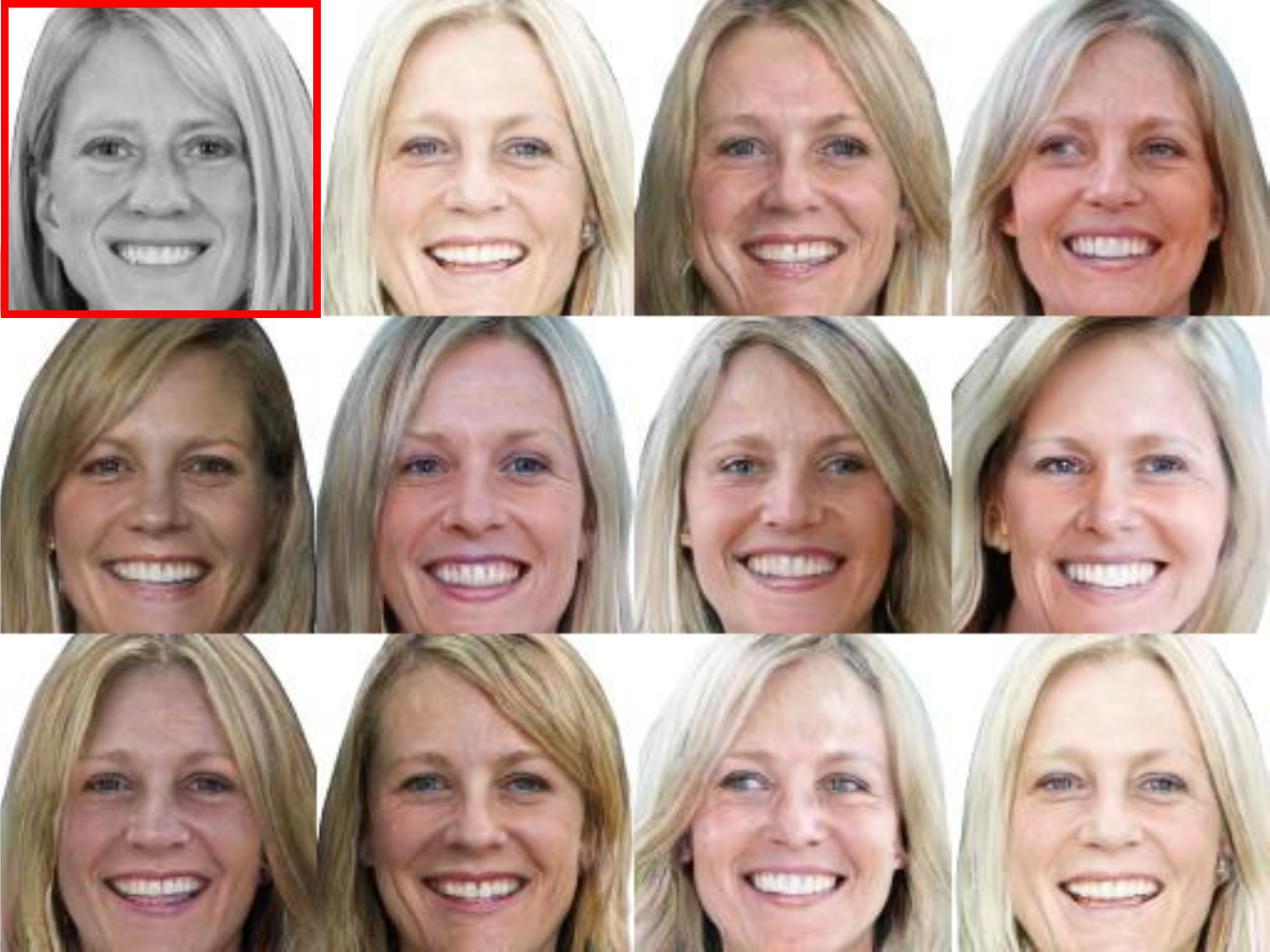} \\
\includegraphics[width=0.45\linewidth]{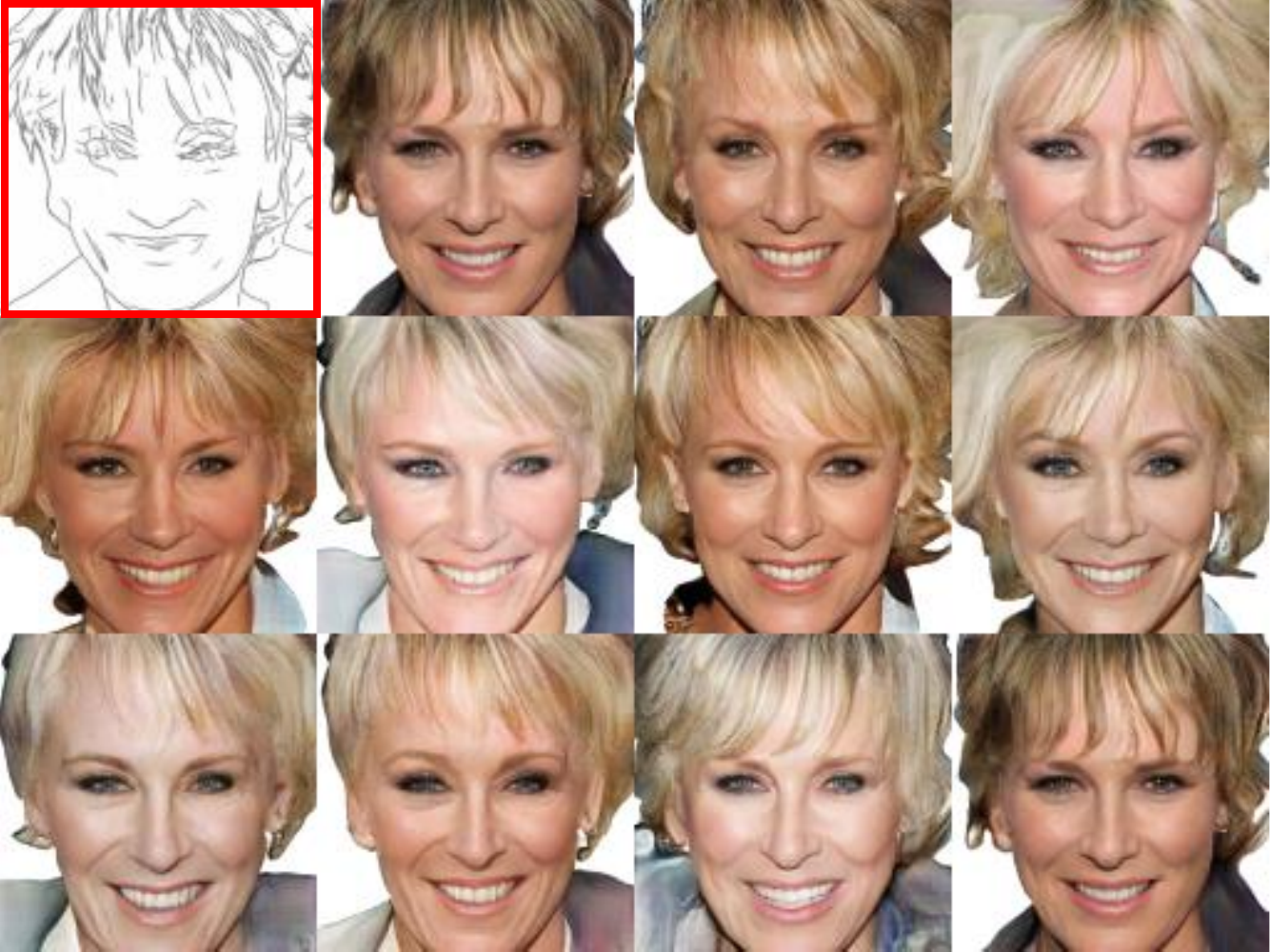} &
\includegraphics[width=0.45\linewidth]{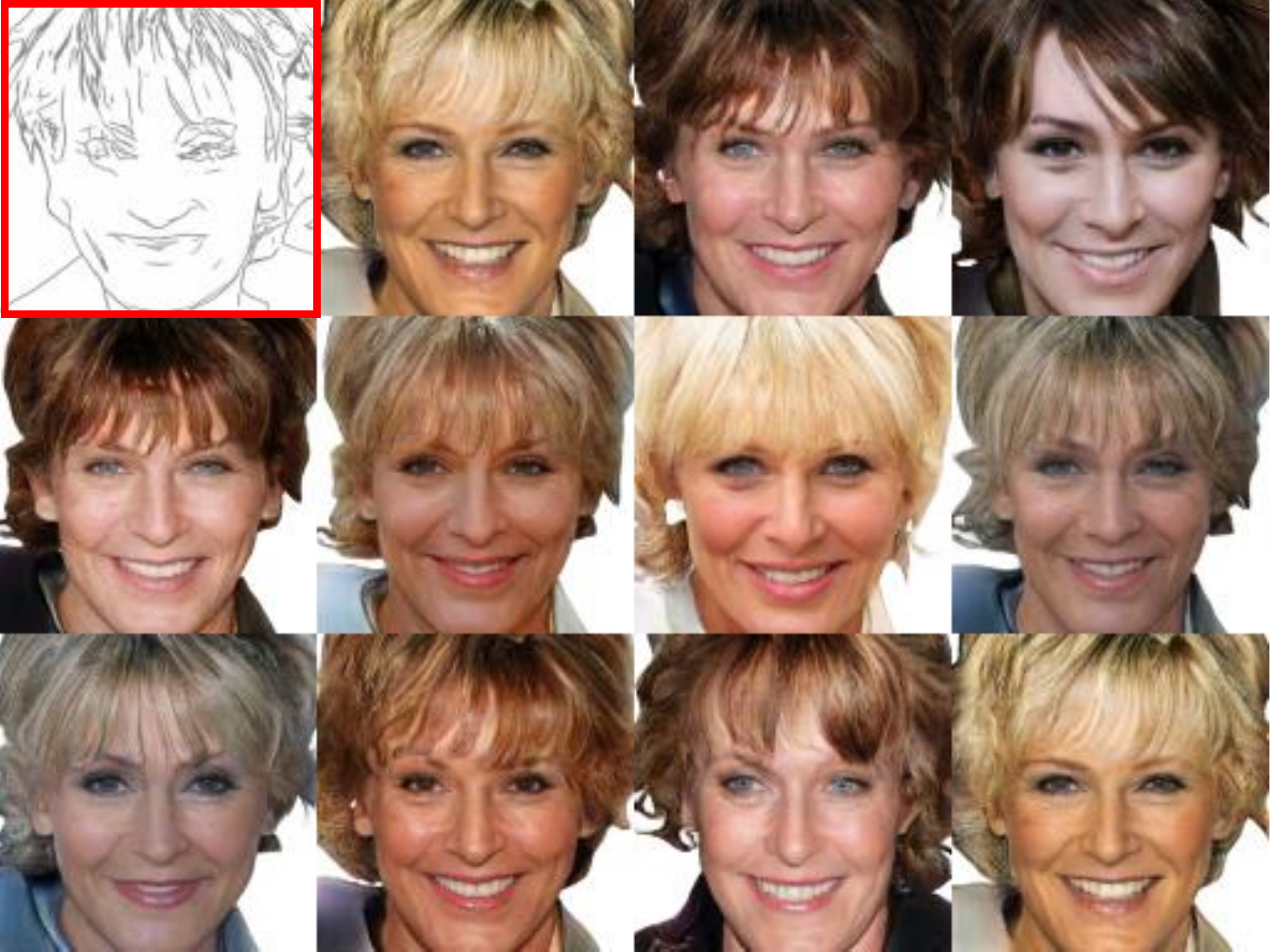} \\

{\footnotesize(a) Trained without PD loss }& {\footnotesize (b) Trained with PD loss }
\end{tabular}
}
\caption{\small Qualitative analysis of the PD loss. Along with condition inputs which are visualized with red rectangles (grayscale in row 1, sketch in row 2), Eleven output images generated with different noise codes are visualized.}
\label{fig:quali_div}
\end{figure}

\begin{table}[t]
\centering
\renewcommand{\tabcolsep}{1mm}
\renewcommand{\arraystretch}{0.8}
\captionsetup{width=0.9\linewidth}
{\scriptsize
\begin{tabular}{c|c|c|c|c}
\hline
 & \multicolumn{2}{c|}{without PDloss} & \multicolumn{2}{c}{with PDloss} \\ \hline
Condition Types & Precision↑ & Recall↑ & Precision↑ & Recall↑ \\ \hline
Color Image & 0.899 & 0.520 & 0.900 & 0.550 \\ \hline
Grayscale & 0.897 & 0.532 & 0.900 & 0.536 \\ \hline
Sketch & 0.904 & 0.547 & 0.892 & 0.567 \\ \hline
Low-Resolution & 0.910 & 0.497 & 0.896 & 0.514 \\ \hline
Text & 0.898 & 0.489 & 0.891 & 0.510 \\ \hline
Average & 0.902 & 0.517 & 0.895 & 0.535 \\ \hline
\end{tabular}%
}
\caption{\small Effect of the PD loss on precision and recall for measuring the fidelity and diversity, respectively, on the CelebA-HQ Dataset.}
\label{tb:precision_recall_per_cond}
\end{table}

\subsubsection{Enhanced Diversity } 

Because the PD loss proposed in this paper can improve the diversity of the generated images, we analyze the effect of the PD loss by taking recall and precision measurements. As shown in \tabref{tb:precision_recall_per_cond}, as a result of applying the PD loss, the recall value is improved by about 3.5\%, and the precision shows a decrease of about 0.77\% on average, showing minimul degradation of visual quality. In addition, the recall is improved in all conditions; in particular, for the color and grayscale condition settings, both precision and recall are improved. From this result, applying the PD loss can increase the diversity while maintaining similar fidelity outcomes. \figref{fig:quali_div} visualizes the result for a qualitative comparison of cases with and without the PD loss. The PD loss encourages the model to generate more diverse images compared to those without this loss, not only on the hair and skin color but also on the illumination.




\subsubsection{Pose Penalty}
To validate the importance of the pose-penalty in relation to the diversity-sensitive loss ~\cite{yang2019diversity} for our method, we conduct an ablation study to confirm the effect of the pose-penalty when attaching the diversity-sensitive loss when training. As shown in \figref{fig:abl-pose} (a), the diversity-sensitive loss alone prevents the network from properly learning the canonical views of objects. This implies that the model maximizes the pixel-level difference causing the pose difference of the output image, which is an undesirable effect. 
With the PD loss, the network properly learns to maximize the style difference while maintaining the pose. For a quantitative validation, we measure the head poses of randomly generated canonical view images using the pre-trained head pose estimator ~\cite{yang2019fsa}. As shown in~\figref{fig:abl-pose} (b), view-consistency is maintained with a pose-penalty by a large margin compared to the result without a pose-penalty, by showing the lower standard deviation of angles of identical view images. Note that the difference in the standard deviation of the rotation angle is larger than that in the elevation angle, as the prior camera pose distribution has a broader range of the rotation angle. 


\begin{figure}[t]
\captionsetup{width=0.9\linewidth}
\begin{center}
\begin{tabular}{@{}c}
\includegraphics[width=0.8\linewidth]{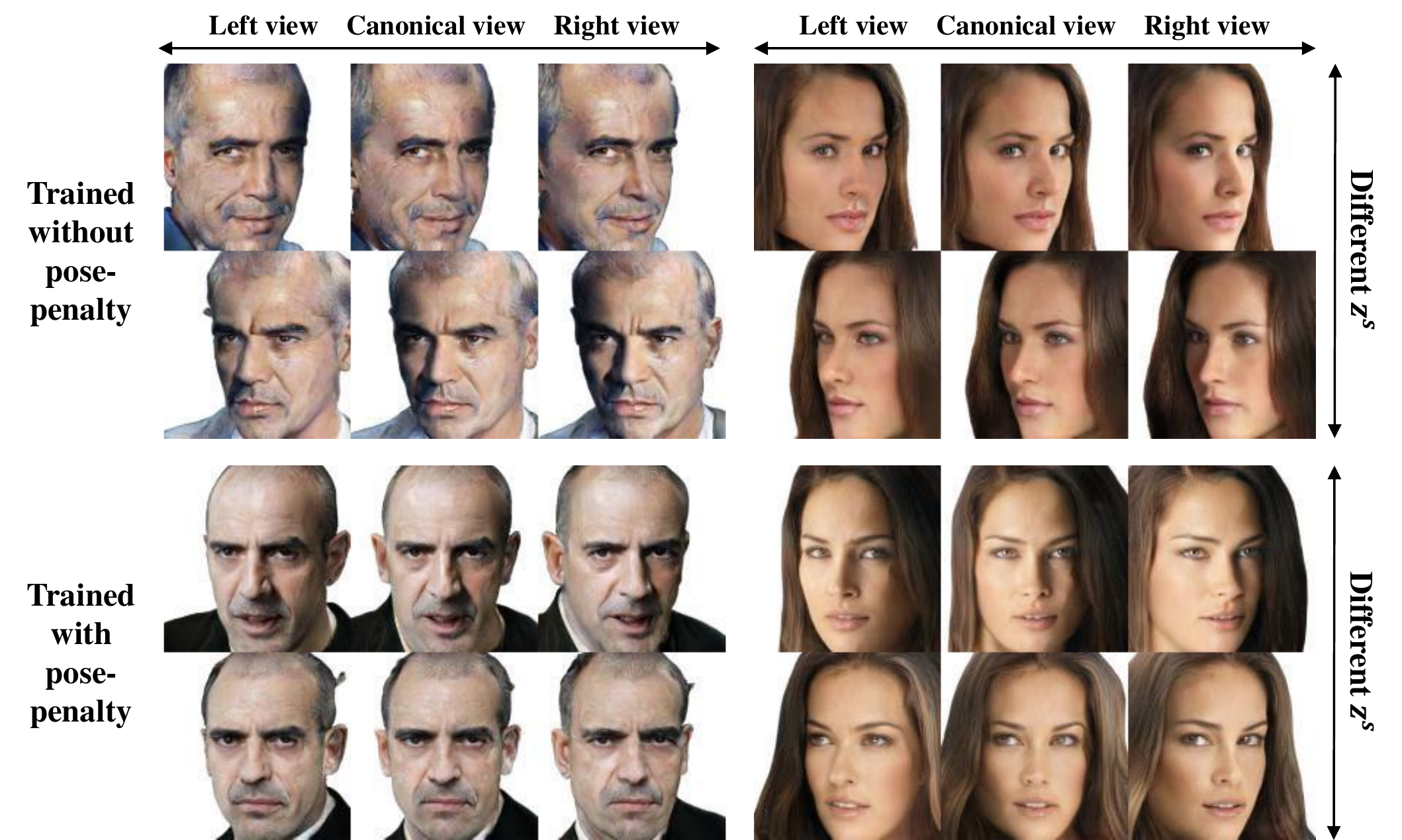} \\
{\small (a) Trained without and with a pose-penalty} \\
\includegraphics[width=0.7\linewidth]{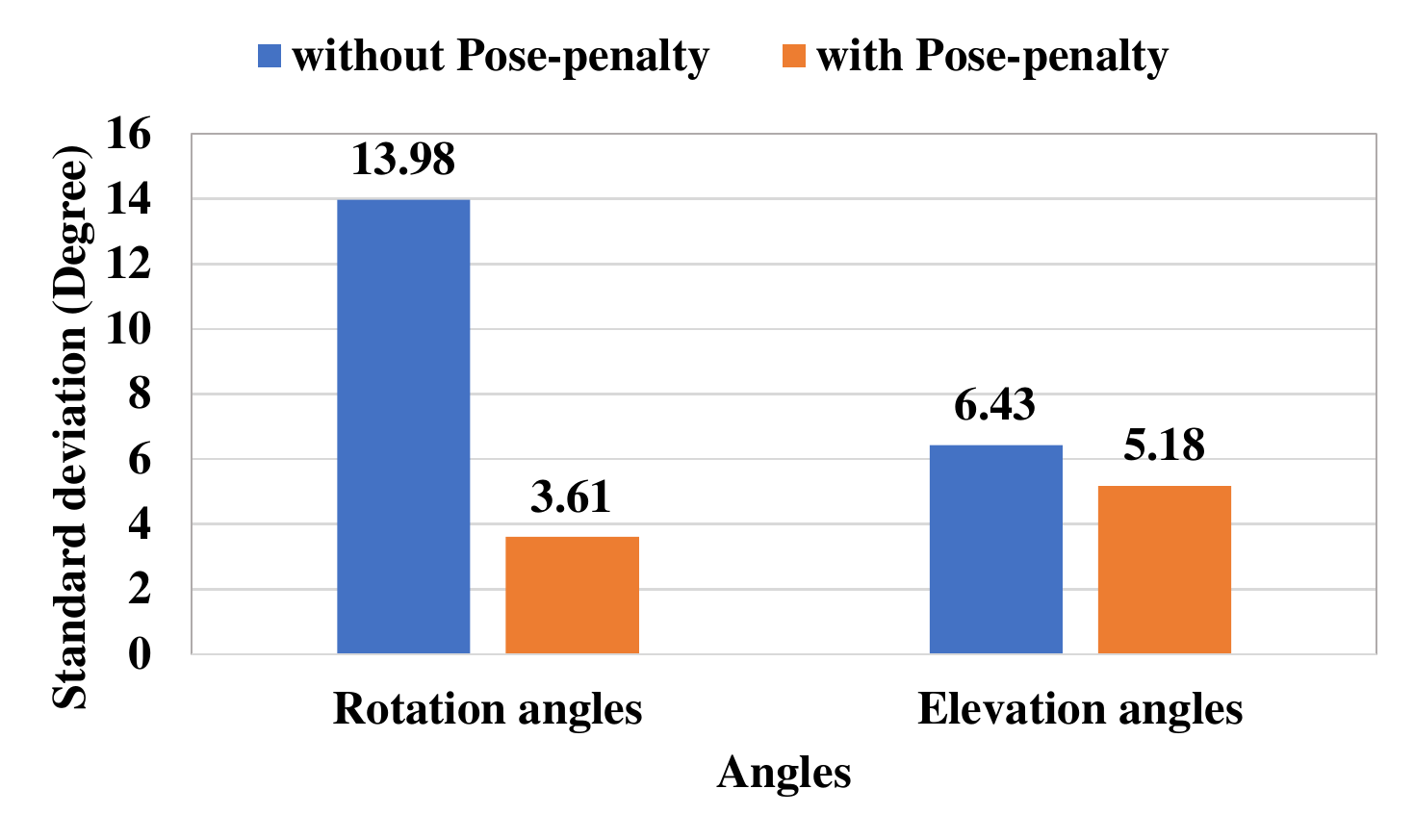} \\
{\small (b) Standard deviations of head poses.}
\end{tabular}
\caption{\small Effects of the pose-penalty when attaching the diversity-sensitive loss when training. As shown in (a), for the result trained without a pose-penalty, the canonical view varies as different shape noise codes are sampled. In contrast, the result trained with a pose-penalty maintains the canonical view with different shape noise codes. 
(b) shows the standard deviation of head poses of randomly generated canonical view images.}
\label{fig:abl-pose}
\end{center}
\end{figure}

\subsubsection{Results of CUB-200} 
\figref{fig:quali_cub} shows qualitative results on the CUB-200 dataset for text input condition. Our proposed model successfully utilizes contextual information in the given text input to generate conditional multi-view images. However, for most existing NeRF-based generative models, we empirically find that the visual quality is degraded for CUB-200 dataset in certain range of viewpoints. We suppose the performance degradation comes from large discrepancy between the prior camera pose distribution and the real one, as described in \cite{niemeyer2021campari}. We plan to address this issue for future work.

\begin{figure}[h!]
\captionsetup{width=0.9\linewidth}
\centering
\begin{tabular}{@{}c@{\hskip 0.02\linewidth}c}
\includegraphics[width=0.3\linewidth]{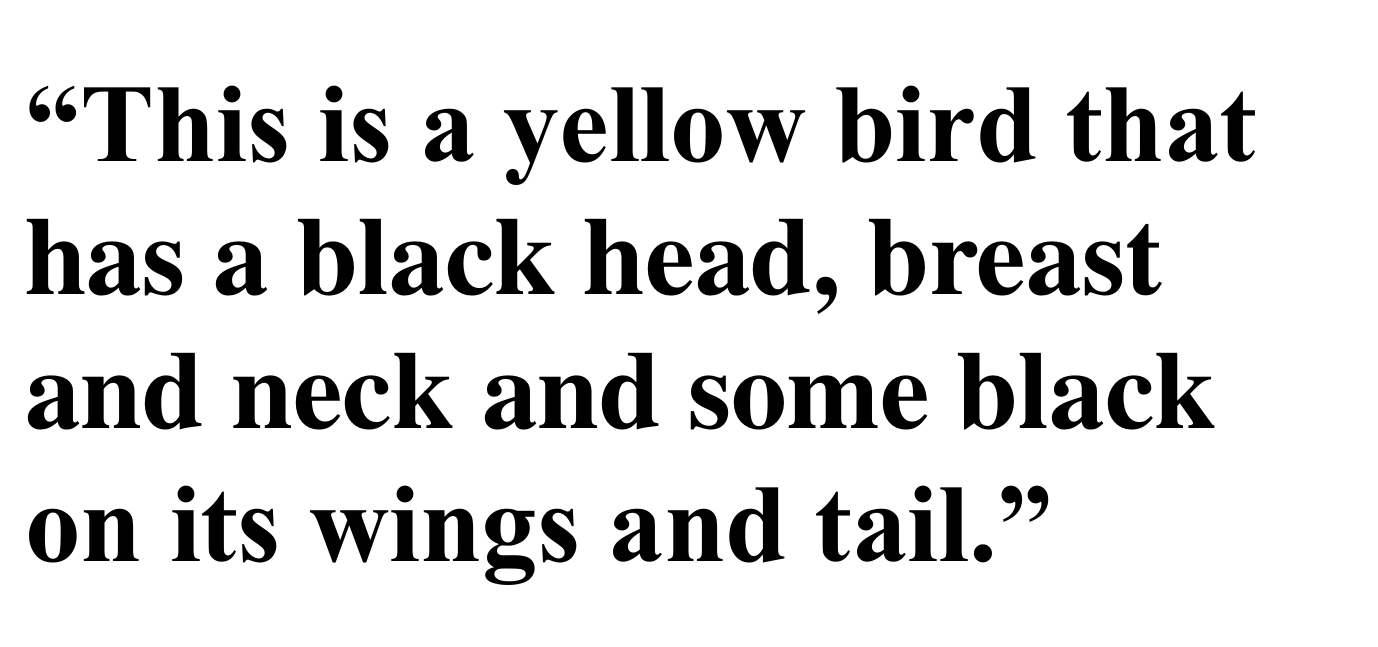} &
\includegraphics[width=0.53\linewidth]{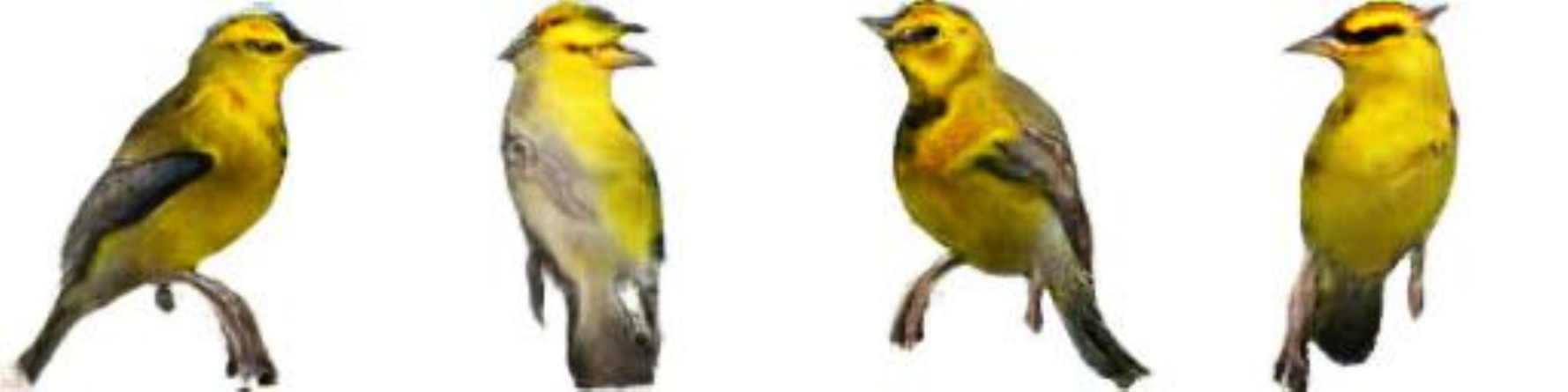} \\
\includegraphics[width=0.3\linewidth]{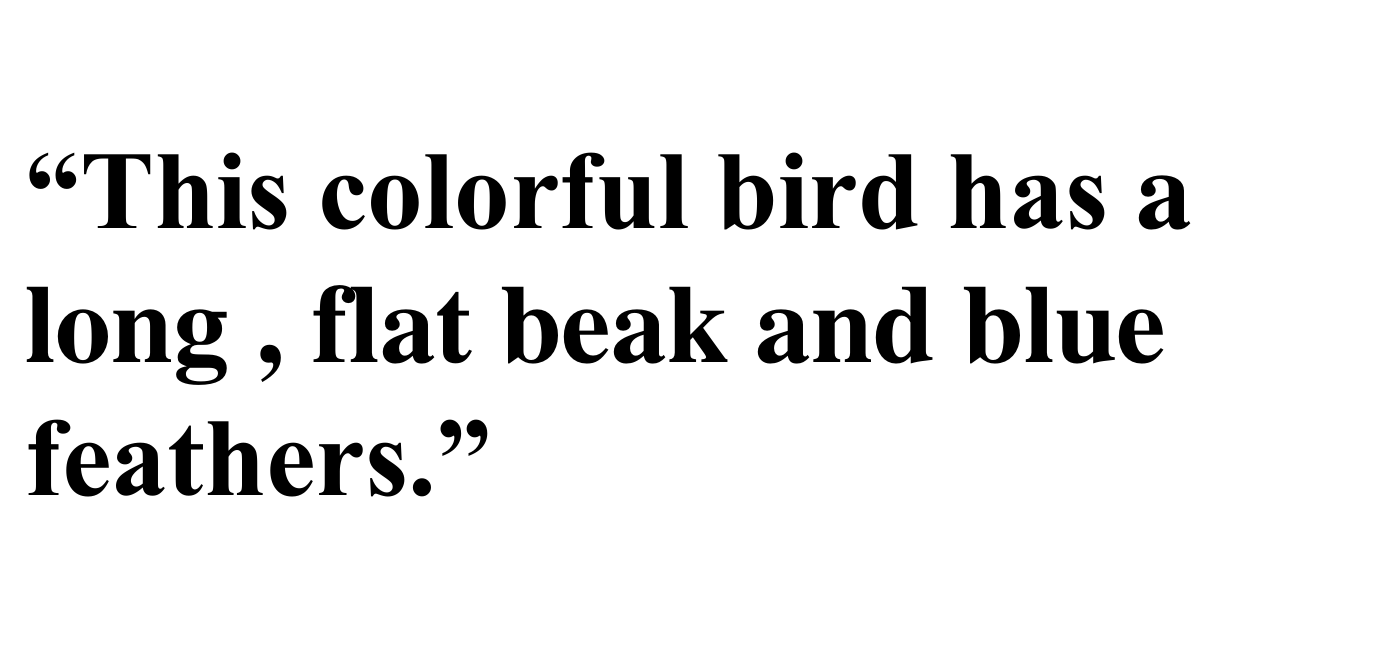} &
\includegraphics[width=0.53\linewidth]{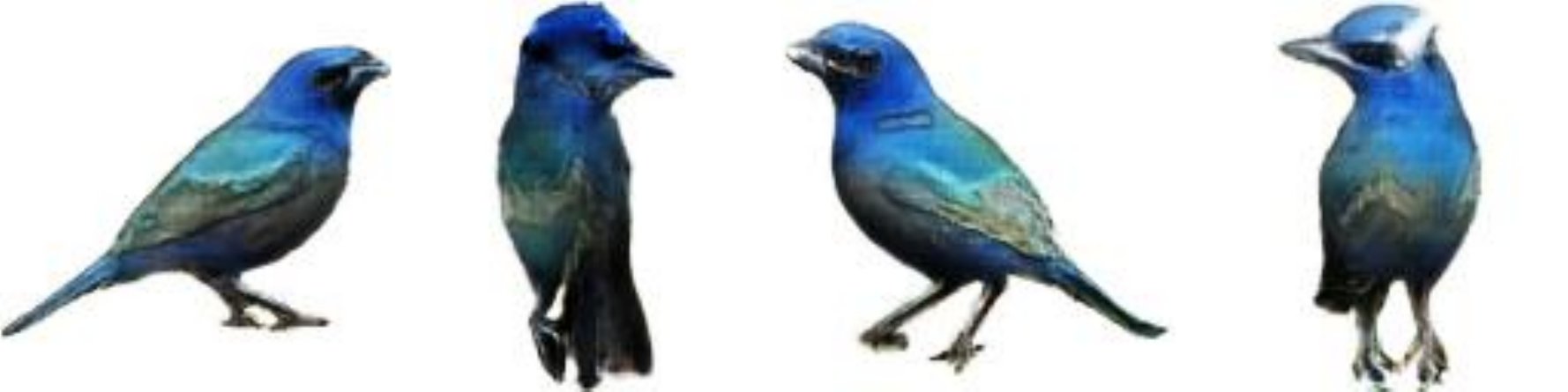} \\
{\footnotesize Input text }& {\footnotesize Multi-view output images }
\end{tabular}
\caption{\small  Multi-view output images in CUB-200 dataset.}
\label{fig:quali_cub}
\end{figure}

\section{Conclusion}
In this paper, we propose a novel conditional generative model called CG-NeRF, which takes the existing generative NeRF to the next level. CG-NeRF creates photorealistic view-consistent images reflecting the condition input, such as sketches or text. Our framework effectively extracts both the shape and appearance from the condition and generates diverse images by adding details through noise codes. In addition, we propose the PD loss to enhance the variety of generated images while maintaining view consistency. Experimental results demonstrate that our method achieves state-of-the-art performance qualitatively and quantitatively based on the quality metrics of FID, precision, and recall. In addition, the proposed method generates various images reflecting the properties of the condition types in terms of the shape and appearance. 
\end{abstract}

\newpage
\clearpage
\appendix

\bibliography{aaai22}


\end{document}